\documentclass[namedate,webpdf,modern,large]{oup-authoring-template}

\usepackage{amsmath}
\usepackage{natbib,xr}

\usepackage{hyperref}
\hypersetup{
  colorlinks=true,        
  citecolor=blue,         
  linkcolor=blue,        
  urlcolor=blue           
}

\graphicspath{{Fig/}}

\theoremstyle{thmstyleone}%
\theoremstyle{thmstyletwo}%
\theoremstyle{thmstylethree}%

\newcommand{\bd}{\boldsymbol} 
\newcommand{\mb}{\mathbf} 

\newcommand{\be}{\begin{equation}}
\newcommand{\ee}{\end{equation}}

\DeclareMathOperator*{\argmin}{arg\,min}
\DeclareMathOperator*{\argmax}{arg\,max}
\DeclareMathOperator{\cov}{cov}
\DeclareMathOperator{\corr}{corr}
\DeclareMathOperator{\var}{var}

\DeclareMathOperator{\st}{\text{s.t.}}
\DeclareMathOperator{\tr}{tr}

\usepackage{algorithm}
\usepackage{algpseudocode}
\makeatletter
\renewcommand{\fnum@algorithm}{\fname@algorithm~\thealgorithm.}
\makeatother

\algnewcommand\INPUT{\item[\textbf{Input:}]}%
\algnewcommand\OUTPUT{\item[\textbf{Output:}]}%

\usepackage{subcaption}

\begin{document}

\journaltitle{Journal Title Here}
\DOI{DOI HERE}
\copyrightyear{2022}
\pubyear{2019}
\access{Advance Access Publication Date: Day Month Year}
\appnotes{Paper}

\firstpage{1}


\title[SNGCCA]{Nonlinear Sparse Generalized Canonical Correlation Analysis for Multi-view High-dimensional Data}

\author[1,3]{Rong Wu}
\author[4]{Ziqi Chen}
\author[5]{Gen Li}
\author[\ORCID{0000-0002-6968-4063},1,2,$\ast$]{Hai Shu}

\authormark{Rong Wu et al.}

\address[1]{\orgdiv{Department of Biostatistics, School of Global Public Health}, \orgname{New York University}, \orgaddress{New York, \state{NY}, \country{USA}}}

\address[2]{\orgdiv{Center for Health Data Science, School of Global Public Health}, \orgname{New York University}, \orgaddress{New York, \state{NY}, \country{USA}}}

\address[3]{\orgdiv{Department of Epidemiology and Biostatistics}, \orgname{University of California, San Francisco}, \orgaddress{\state{CA}, \country{USA}}}

\address[4]{\orgdiv{School of Statistics, KLATASDS-MOE},
\orgname{East China Normal University}, \orgaddress{Shanghai,  \country{China}}}

\address[5]{\orgdiv{Department of Biostatistics}, \orgname{University of Michigan}, \orgaddress{Ann Arbor, \state{MI}, \country{USA}}}

\corresp[$\ast$]{Corresponding author. Email: \href{hs120@nyu.edu}{hs120@nyu.edu}}




\abstract{\textbf{Motivation:} 
Biomedical studies increasingly produce multi-view high-dimensional datasets (e.g., multi-omics) that demand integrative analysis. 
Existing canonical correlation analysis (CCA) and generalized CCA  methods address at most two of the following three key aspects simultaneously: (i) nonlinear dependence, (ii) sparsity for variable selection, and (iii) generalization to more than two data views.
 There is a pressing need for CCA methods that integrate all three aspects to effectively analyze multi-view high-dimensional data.
\\
\textbf{Results:} 
We propose three nonlinear, sparse, generalized CCA methods, HSIC-SGCCA, SA-KGCCA, and TS-KGCCA, for variable selection in multi-view high-dimensional data. 
These methods extend existing SCCA-HSIC, SA-KCCA, and TS-KCCA from two-view to multi-view settings. While SA-KGCCA and TS-KGCCA yield multi-convex optimization problems solved via block coordinate descent, HSIC-SGCCA introduces a necessary unit-variance constraint previously ignored in SCCA-HSIC, resulting in a nonconvex, non-multiconvex problem.
We efficiently address this challenge by integrating the block prox-linear method with the  linearized
alternating direction method of multipliers. 
Simulations and TCGA-BRCA data analysis demonstrate that HSIC-SGCCA outperforms competing methods in multi-view variable selection.
\\
\textbf{Availability and implementation:}  Code is available at \url{https://github.com/Rows21/NSGCCA}.\\
}


\maketitle

\section{Introduction}\label{sec: intro}
Modern biomedical studies often collect multi-view data, consisting of multiple data types (i.e., “views”) measured on the same set of  objects. Each data view provides a unique but complementary perspective on the underlying biological processes or disease mechanisms. For instance, The Cancer Genome Atlas (TCGA) project (\url{www.cancer.gov/tcga}) systematically profiles various tumor types, such as breast carcinoma, by collecting multi-omics data, including mRNA expression, DNA methylation, and miRNA expression,
to characterize genomic alterations and identify potential biomarkers.
Integrating multi-view data can offer a more holistic understanding of complex diseases,  enhancing  diagnosis, prediction, risk stratification, and the discovery of novel therapeutic targets \citep{chen2023applications}. However, combining these heterogeneous data views poses significant analytical challenges, such as high dimensionality and complex relationships between data views. Consequently, there is growing interest in developing powerful statistical and machine learning methods to fully exploit the potential of multi-view data
\citep{yu2025review}.

A representative class  of multi-view 
learning methods is 
canonical correlation analysis (CCA) and its various extensions.
CCA \citep{hotelling1992relations}
is a classical two-view method
that evaluates the linear association
between two data views
by identifying
their most correlated 
linear transformations. 
In each view’s linear transformation, the coefficients of the variables reflect their respective contributions to establishing that view’s linear association with the other view.
To extend CCA to more than two views,
various generalized CCA (GCCA) methods \citep{horst1961generalized, carroll1968generalization, kettenring1971canonical}
have been proposed
to identify the linear transformations of  views that maximize the overall linear association,
defined by different optimization criteria.

However, for high-dimensional data, 
the empirical estimation of these CCA and GCCA methods 
becomes inconsistent when
sample covariance matrices are used to estimate the true covariance matrices,
due to the accumulation of estimation errors
over matrix entries \citep{MR950344}. 
To overcome the curse of high dimensionality,
many sparse CCA
\citep[SCCA;][]{waaijenborg2008,parkhomenko2009sparse,witten2009penalized,hardoon2011sparse,
chu2013sparse,
gao2017sparse,
mai2019iterative, gu2020sparse,lindenbaum2021l0,li2024on}
and sparse GCCA
  \citep[SGCCA;][]{witten2009extensions,tenenhaus2014variable,fu2017scalable,kanatsoulis2018structured,rodosthenous2020,9619966,lv2024sparse} 
methods have been developed.
These methods reduce the variable dimension for estimation by
imposing sparsity constraints on the 
coefficients of the variables in each view’s linear transformation,
using various penalties, optimization criteria, and algorithms.
The imposed sparsity naturally leads to variable selection, enabling more effective downstream analysis and improved interpretation.
Variables with nonzero coefficients 
are selected as  low-dimensional representatives
of each  view, as they  retain the linear transformations that maximize the overall linear association between views.

Moreover, to assess the nonlinear association 
between data views, various nonlinear extensions of CCA and GCCA have been devised.
Kernel CCA  \citep[KCCA;][]{bach2002kernel,Fukumizu2007} and kernel GCCA \citep[KGCCA;][]{tenenhaus2015kernel}
measure the nonlinear association  by
identifying
the most correlated
nonlinear transformations of views
within reproducing kernel Hilbert spaces \citep[RKHSs;][]{aronszajn1950theory}.
An RKHS with a Gaussian kernel provides an accurate approximations of the space of finite-variance functions,  serving as a manageable surrogate that simplifies computation and analysis \citep{steinwart2008support}.
Alternatively, 
deep (G)CCA \citep[D(G)CCA;][]{andrew2013deep,benton2017deep} and  variants \citep{wang2015deep,li2020application,DCCA-SCO}
model the most correlated nonlinear transformations
by deep neural networks (DNNs),
leveraging their high expressive power to approximate 
any continuous functions \citep{gripenberg2003approximation}.
Instead of using  Pearson correlation, 
a linear dependence measure,
between nonlinear transformations of views,
hsicCCA \citep{{chang2013canonical}}
maximizes a nonlinear dependence measure,  
the Hilbert-Schmidt independence criterion \citep[HSIC;][]{gretton2005measuring},
between  linear transformations~of~views.

Unlike linear (G)CCA,  
applying sparse constraints for variable selection in K(G)CCA and D(G)CCA methods is not straightforward, as their nonlinear transformations
of views do not have coefficients that correspond one-to-one with individual variables.
To address this,
\citet{SAFKCCA} propose sparse additive KCCA (SA-KCCA), which assumes that each view's nonlinear transformation is a sparse additive function of individual variables, while
\citet{yoshida2017sparse} introduce two-stage KCCA (TS-KCCA) using sparse multiple kernel learning.
 \citet{lindenbaum2021l0}
propose  $\ell_0$-DCCA, which induces sparsity 
by applying stochastic gates to individual variables 
and penalizing the DCCA objective with the mean 
$\ell_0$ norm of these gates.
In contrast, 
SCCA-HSIC \citep{uurtio2018sparse} enforces sparsity by penalizing hsicCCA  
with the $\ell_1$ penalty 
on the coefficients of variables in each view's linear transformation. 

Nonetheless, these nonlinear SCCA methods
are  limited to two-view data and cannot be directly applied  to multi-view data with more than two views, as applying them to each pair of views fails to produce identical  transformations within each individual view.
To the best of our knowledge, a 
nonlinear SGCCA method
has not yet  been developed in the existing literature.
 In this paper, we present the first detailed formulation and implementation of nonlinear SGCCA methods for multi-view high-dimensional data.

We propose three nonlinear 
SGCCA methods, 
HSIC-SGCCA, SA-KGCCA, and TS-KGCCA,
for variable selection in multi-view high-dimensional data.
These new methods extend SCCA-HSIC, SA-KCCA,
and TS-KCCA to more than two views using optimization criteria 
similar to the SUMCOR criterion~\citep{kettenring1971canonical}.
For HSIC-SGCCA, we  incorporate a unit-variance constraint, which is necessary but ignored
in SCCA-HSIC (see Section~\ref{sec: SCCA-HSIC}).
To solve
the challenging optimization problem of HSIC-SGCCA, which is neither convex nor multi-convex,
we propose an efficient algorithm 
that integrates 
the  block prox-linear (BPL) method
\citep{xu2017globally}
and the linearized alternating direction method of multipliers (ADMM) 
\citep[LADMM;][]{fang2015generalized}.
The optimization problems for SA-KGCCA and TS-KGCCA
are multi-convex, and we solve them using
the block coordinate descent (BCD) strategy \citep{MR3444832}.
We compare the proposed methods against competing approaches in
both simulations and a real-data analysis on TCGA-BRCA, a multi-view dataset for breast invasive carcinoma from TCGA \citep{Kobo12}. 
The proposed HSIC-SGCCA achieves the best performance in variable selection in simulations, and its selected variables excel in breast cancer subtype separation and survival time prediction in the TCGA-BRCA data analysis.

\section{Preliminaries}\label{sec: Preliminaries}

\subsection{Notation}
In this paper, we
consider  multi-view
data with
$K\ge 2$ data views
measured on the same set of $n$ objects.
Each $k$-th data view consists of $p_k$ random variables.
We denote
the multi-view data
as $\{\bd{x}_1^{(i)},\dots, \bd{x}_K^{(i)}\}_{i=1}^n$,
where $\bd{x}_k^{(i)}\in \mathbb{R}^{p_k}$
is the $k$-th data view
for the $i$-th object.
Assume that 
$\{\bd{x}_1^{(i)},\dots, \bd{x}_K^{(i)}\}_{i=1}^n$
are $n$
independent and identically distributed
(i.i.d.) observations
of the random vectors $\{\bd{x}_1,\dots, \bd{x}_K\}$.
Let $\bd{x}_k^{[j]}$ denote
the $j$-th entry of~$\bd{x}_k$.

\subsection{RKHS}\label{sec: RKHS}

A Hilbert space $\mathcal{H}$ of 
functions from a non-empty  set $\mathcal{X}$
to $\mathbb{R}$
 is called
a RKHS if it has a 
reproducing kernel
$\kappa$, defined as  a
function from $\mathcal{X}\times \mathcal{X}$ to $\mathbb{R}$ 
that satisfies
$\kappa(\cdot, x)\in \mathcal{H}$
and $f(x)=\langle f,\kappa(\cdot, x)\rangle_\mathcal{H}$ for all $x\in \mathcal{X}$ and $f\in \mathcal{H}$
\citep{steinwart2008support,aronszajn1950theory}.
Here, 
$\langle\cdot,\cdot\rangle_{\mathcal{H}}$
is the inner product in $\mathcal{H}$,
and
its induced norm is
denoted as
$\|\cdot \|_{\mathcal{H}}$.
The RKHS $\mathcal{H}$ can be written as 
the closure of 
the linear span of the functions
$\{\kappa(\cdot, x):x\in\mathcal{X}\}$:
\begin{align*}
\mathcal{H}&=\overline{\text{span}
(\{\kappa(\cdot, x):x\in\mathcal{X}\})}\\
&=\overline{\left\{\sum_{i=1}^m \alpha_i \kappa(\cdot, x_i): m\in \mathbb{Z}^+, \alpha_i\in \mathbb{R}, x_i\in \mathcal{X}  \right\}}.
\end{align*}

The real-valued RKHS $\mathcal{H}_\sigma$ on $\mathcal{X}=\mathbb{R}^d$ with
a Gaussian kernel $\kappa_\sigma$ provides  an accurate approximation of 
$L^2(\text{P})$ 
for any probability distribution $\text{P}$ on $\mathbb{R}^d$,
and is thus used as
a manageable surrogate for $L^2(\text{P})$
to simplify computation and analysis.
The Gaussian kernel is defined as
 $\kappa_\sigma(\bd{x},\bd{x}')=\exp\{-\|\bd{x}-\bd{x}'\|_2^2/(2\sigma^2)\}$ for $\bd{x},\bd{x}'\in\mathbb{R}^d$ and $\sigma>0$,
and $L^2(\text{P})$ is the space of all functions $f$ with 
$\var(f(\bd{x}))<\infty$
for $\bd{x}$ following $\text{P}$.
The  effectiveness of $\mathcal{H}_\sigma$ in approximating  $L^2(\text{P})$ stems from the fact
that 
$\mathcal{H}_\sigma$ is
dense in $L^2(\text{P})$ 
for any $\sigma>0$  and  $\text{P}$ on $\mathbb{R}^d$
\citep{steinwart2008support}. That is,
for any $\epsilon>0$ and $g\in L^2(\text{P})$,
there exists an $f\in \mathcal{H}_\sigma$
such that $\|f-g\|_{L^2(\text{P})}^2=\int_{\mathbb{R}^d} |f(\bd{x})-g(\bd{x})|^2  d\text{P}(\bd{x})< \epsilon$.
Thus, 
$\mathcal{H}_\sigma$ 
can accurately 
approximate 
$L^2(\text{P})$, 
regardless of the value of the Gaussian kernel bandwidth~$\sigma$.
In practice,
for observations of $\bd{x}$,
$\sigma^2$ is often
set to
the trace of the sample covariance matrix 
\citep{ramdas2015adaptivity,chen2024high}
or the median of squared Euclidean distances between 
observations
\citep{tenenhaus2015kernel}.

\subsection{HSIC
}\label{subsec41}

HSIC was first introduced by \citet{gretton2005measuring} 
to measure the dependence between two random vectors $\bd{x}_1\in\mathbb{R}^{p_1}$ and $\bd{x}_2\in\mathbb{R}^{p_2}$, which is defined
as the squared Hilbert–Schmidt norm of their
cross-covariance operator 
between their respective RKHSs $\mathcal{H}_1$ and $\mathcal{H}_2$.
Equivalently, yet more intuitively,  HSIC can be written as 
the sum of squared kernel constrained  covariances
\citep{chen2021estimations}:
\[
\mathrm{HSIC}(\bd{x}_1,\bd{x}_2;\mathcal{H}_1,\mathcal{H}_2)
=\sum_{j=1}^\infty \cov^2(\phi_{1j}(\bd{x}_1),\phi_{2j}(\bd{x}_2)),
\]
\vspace{-0.5cm}
\begin{align*}
\text{where}\quad\quad &
\{\phi_{kj}\}_{k=1}^2 
=\argmax_{\{f_{kj}\in\mathcal{H}_k\}_{k=1}^2} \cov(f_{1j}(\bd{x}_1),f_{2j}(\bd{x}_2))\nonumber\\
&\st~ \|f_{kj}\|_{\mathcal{H}_k}=1,\quad
\langle f_{kj}, \phi_{ki}  \rangle_{\mathcal{H}_k}
=0,\quad
1\le i\le j-1.\nonumber
\end{align*}
For ease of estimation, 
it is written 
with kernels $\kappa_1,\kappa_2$ of $\mathcal{H}_1,\mathcal{H}_2$~as
\begin{align}
&\mathrm{HSIC}(\bd{x}_1,\bd{x}_2;\mathcal{H}_1,\mathcal{H}_2)
\nonumber\\
&= \mathrm{E}[\kappa_1(\bd{x}_1,\bd{x}_1^{\prime})\kappa_2(\bd{x}_2,\bd{x}_2^{\prime})] + \mathrm{E}[\kappa_1(\bd{x}_1,\bd{x}_1^{\prime})]\mathrm{E}[\kappa_2(\bd{x}_2,\bd{x}_2^{\prime})] \nonumber\\ 
&\quad 
-2\mathrm{E}\big[\mathrm{E}[\kappa_1(\bd{x}_1,\bd{x}_1^{\prime})|\bd{x}_1]\mathrm{E}[\kappa_2(\bd{x}_2,\bd{x}_2^{\prime})|\bd{x}_2]\big],
\label{eqn: HSIC in kernel}
\end{align}
where $\{\bd{x}_1',\bd{x}_2'\}$ is an i.i.d. copy of $\{\bd{x}_1,\bd{x}_2\}$.
Replacing the means with sample means
and reorganizing
yields
an consistent estimator of HSIC,
known as the empirical HSIC \citep{GrettonKIT}:
\be\label{eqn: empirical HSIC}
\widehat{\text{HSIC}}(\{\bd{x}_1^{(i)},\bd{x}_2^{(i)}\}_{i=1}^n;\mathcal{H}_1,\mathcal{H}_2)
=\frac{\tr(\mb{K}_1\mb{H}\mb{K}_2\mb{H})}{n^2},
\ee
where $\mb{K}_k=[\kappa_k(\bd{x}_k^{(i)},\bd{x}_k^{(j)})]_{1\le i,j\le n}\in\mathbb{R}^{n\times n}$, 
$\mb{H}=\mb{I}_n-\bd{1}_n\bd{1}_n^\top/n$,
$\mb{I}_n$ is the $n\times n$ identity matrix,
and $\bd{1}_n$ is the $n\times 1$ vector of ones.

\subsection{(G)CCA and S(G)CCA}
CCA \citep{hotelling1992relations} is a classic approach to studying the linear association between two data views $\bd{x}_1\in\mathbb{R}^{p_1}$ and $\bd{x}_2\in\mathbb{R}^{p_2}$. 
CCA seeks  canonical coefficient vectors $\bd{u}_1\in\mathbb{R}^{p_1}$ and $\bd{u}_2\in\mathbb{R}^{p_2}$ that maximizes
the correlation between 
$\bd{u}_1^\top\bd{x}_1$ and $\bd{u}_2^\top\bd{x}_2$:
$$
       \max_{\{\bd{u}_k\in \mathbb{R}
     ^{p_k}\}_{k=1}^2} 
     \cov(\bd{u}_1^\top\bd{x}_1,\bd{u}_2^\top\bd{x}_2)  
 \quad       \st      \quad \var(\bd{u}_k^\top\bd{x}_k)      
        =1.
$$

GCCA methods \citep{horst1961generalized, carroll1968generalization, kettenring1971canonical}
extend CCA to $K\ge 2$
data views $\{\bd{x}_k\in \mathbb{R}^{p_k}\}_{k=1}^K$
 with different optimization criteria.
 Two main
 criteria are 
SUMCOR and MAXVAR \citep{kettenring1971canonical}. 
The SUMCOR criterion maximizes 
the sum of pairwise correlations
between~$\{\bd{u}_k^\top\bd{x}_k\}_{k=1}^K$:
\be\label{obj: SUMCOR}\hspace{-0.18cm}
       \max_{\{\bd{u}_k\in \mathbb{R}
     ^{p_k}\}_{k=1}^K} 
     \sum_{1\le s<t\le K}\cov(\bd{u}_s^\top\bd{x}_s,\bd{u}_t^\top\bd{x}_t)  
~~   \st ~    \var(\bd{u}_k^\top\bd{x}_k)      
        =1.
\ee
Alternatively, the MAXVAR criterion
maximizes the variance of the first principal component of $\{\bd{u}_k^\top\bd{x}_k\}_{k=1}^K$, 
i.e., the largest eigenvalue
of the covariance matrix
of $(\bd{u}_1^\top\bd{x}_1,\dots, \bd{u}_K^\top\bd{x}_K)^\top$,
which is equivalent to
minimizing
the sum of
mean squared differences
between each 
 $\bd{u}_k^\top(\bd{x}_k-\text{E}[\bd{x}_k])$
and a consensus variable $g$:
\begin{align}\label{obj: MAXVAR}
&\min_{\{\bd{u}_k\in \mathbb{R}
     ^{p_k}\}_{k=1}^K, g\in \mathbb{R}} 
\sum_{k=1}^K\text{E}\left|\bd{u}_k^\top(\bd{x}_k-\text{E}[\bd{x}_k])-g\right|^2 
\\
&\qquad \st~ \var(\bd{u}_k^\top\bd{x}_k)=\var(g)=1,~
\text{E}(g)=0.
\nonumber
\end{align}

For CCA and GCCA,
the covariance matrix  
$\cov(\bd{x}_k,\bd{x}_\ell)$
in their
$
\cov(\bd{u}_k^\top\bd{x}_k,\bd{u}_\ell^\top\bd{x}_\ell)  
=\bd{u}_k^\top
\cov(\bd{x}_k,\bd{x}_\ell) \bd{u}_\ell
$ ($1\le k\le \ell\le K$)
is traditionally estimated by the
sample covariance matrix.
However, for high-dimensional data 
with $n=O(p_k)$,
the sample covariance matrix
is not a consistent estimator of the true covariance matrix \citep{MR950344}
due to  the accumulation of estimation errors over matrix entries.
To overcome the  curse
of high dimensionality,
SCCA 
and 
SGCCA methods (see articles cited in Section~\ref{sec: intro}, {\color{black}paragraph 3})
impose sparsity constraints on 
 canonical coefficient vectors  $\{\bd{u}_k\}_{k=1}^K$
to reduce the variable dimension, using
various penalties, optimization criteria, and algorithms.

Related work on S(G)CCA, K(G)CCA, and DNN-based (G)CCA is detailed in the Supplementary Material.

\subsection{HSIC-based (S)CCA}\label{sec: SCCA-HSIC}
\citet{chang2013canonical}
propose hsicCCA, a nonlinear  CCA
for two data views $\bd{x}_1\in \mathbb{R}^{p_1}$ and $\bd{x}_2\in \mathbb{R}^{p_2}$
based on HSIC solving:
\[
       \max_{\{\bd{u}_k\in \mathbb{R}^{p_k}\}_{k=1}^2}        
     \mathrm{HSIC}(\bd{u}_1^\top\bd{x}_1,\bd{u}_2^\top\bd{x}_2;\mathcal{H}_1,\mathcal{H}_2)
 \quad \st  \quad \| \bd{u}_k\|_2=1,
\]
where 
 $\mathcal{H}_k$ is 
a real-valued RKHS on $\mathbb{R}$.
For high-dimensional two-view data,
\citet{uurtio2018sparse}
introduce SCCA-HSIC, a sparse variant of hsicCCA adding
the $\ell_1$ penalty 
for sparsity on $\{\bd{u}_k\}_{k=1}^2$:
\[
       \max_{\{\bd{u}_k\in \mathbb{R}^{p_k}\}_{k=1}^2}        
     \mathrm{HSIC}(\bd{u}_1^\top\bd{x}_1,\bd{u}_2^\top\bd{x}_2;\mathcal{H}_1,\mathcal{H}_2)
 \quad \st  \quad \| \bd{u}_k\|_1\le s_k.
\]

However, SCCA-HSIC does not impose any normalization constraint
on $\{\bd{u}_k\}_{k=1}^2$. Such
a normalization constraint
is 
necessary. To see this,
assume that $(\bd{x}_1^\top,\bd{x}_2^\top)$ is jointly Gaussian with zero mean, and
use a univariate Gaussian kernel with bandwidth $\sigma=1$ for both $\mathcal{H}_1$ and $\mathcal{H}_2$.
Denote $\rho=\corr(\bd{u}_1^\top\bd{x}_1, \bd{u}_2^\top\bd{x}_2)$
and $\sigma_k^2=\var(\bd{u}_k^\top\bd{x}_k)$.
Then, we have
\begin{align}\label{eqn: HSIC Gaussian}
    & \mathrm{HSIC}(\bd{u}_1^\top\bd{x}_1,\bd{u}_2^\top\bd{x}_2;\mathcal{H}_1,\mathcal{H}_2)
     =\frac{1}{\sqrt{1+2\sigma_1^2+2\sigma_2^2+4\sigma_1^2\sigma_2^2(1-\rho^2)}}\nonumber\\
    &~\qquad+\frac{1}{\sqrt{(1+2\sigma_1^2)(1+2\sigma_2^2)}}
    -\frac{2}{\sqrt{(1+2\sigma_1^2)(1+2\sigma_2^2)-\sigma_1^2\sigma_2^2\rho^2}}.
\vspace{-0.5cm}\end{align}
Since $\bd{u}_1^\top\bd{x}_1$ and $\bd{u}_2^\top\bd{x}_2$ 
follow a bivariate Gaussian distribution,
their dependence is fully determined by their linear relationship.
Maximizing their
 HSIC is expected to be equivalent to
 maximizing their absolute correlation $|\rho|$ as in  linear CCA (up to a sign change of $\rho$).
 From \eqref{eqn: HSIC Gaussian},
this equivalence can be achieved by 
imposing the normalization constraint $\sigma_k^2=\var(\bd{u}_k^\top\bd{x}_k)=1$ for $k=1,2$.

Moreover, 
SCCA-HSIC employs 
a projected stochastic gradient ascent algorithm
with line search to solve its nonconvex  problem, which is computationally intensive and is challenging to adapt for incorporating the desirable unit-variance constraint.
Both hsicCCA and SCCA-HSIC are  limited to  two-view data.

To address these limitations, we propose HSIC-SGCCA, which generalizes SCCA-HSIC to handle $K \ge 2$ data views. Our method incorporates the unit-variance constraint and leverages an efficient algorithm that  integrates the BPL \citep{xu2017globally} and LADMM \citep{fang2015generalized} methods.

\section{Methods}\label{sec: methods}
We focus on HSIC-SGCCA because it involves a more challenging nonconvex, non-multiconvex optimization problem and demonstrates superior performance 
compared to SA-KGCCA and TS-KGCCA
in our simulations and real-data analysis.
In contrast,
SA-KGCCA and TS-KGCCA 
are natural extensions of  SA-KCCA \citep{SAFKCCA} and 
TS-KCCA \citep{yoshida2017sparse}
to $K\ge 2$ data views, resulting in  
multi-convex
optimization problems solved via BCD.
We provide the details of SA-KGCCA and TS-KGCCA in the Supplementary Material.

\subsection{HSIC-SGCCA Problem Formulation} \label{subsec43}

Our proposed HSIC-SGCCA is
a  sparse SUMCOR-like nonlinear GCCA,
which considers the following optimization  problem:
\vspace{-0.2cm}
\begin{align}
        \max_{\{\bd{u}_k\in \mathbb{R}^{p_k}\}_{k=1}^K} &    
     \sum_{1\le s<t\le K}      
     \mathrm{HSIC}(\bd{u}_s^\top\bd{x}_s,\bd{u}_t^\top\bd{x}_t;\mathcal{H}_s,\mathcal{H}_t)-\sum_{k=1}^K{\lambda_k \| \bd{u}_k\|_1}
      \nonumber\\
  & ~~~~~~   \st\quad 
  \var(\bd{u}_k^\top \bd{x}_k)
=\bd{u}_k^\top \mb{\Sigma}_k \bd{u}_k =1,
  \label{orignal multiview obj}
\vspace{-0.2cm}\end{align}
where 
 $\mathcal{H}_k$ is 
a real-valued RKHS on~$\mathbb{R}$,
$\lambda_k>0$ is a tuning parameter for the $\ell_1$ penalty
on sparsity of $\bd{u}_k$,
and $\mb{\Sigma}_k=\cov(\bd{x}_k)$.

For ease of algorithm development, 
we use
the Gaussian kernel
$\kappa_\sigma(x,y)=\exp\{-|x-y|^2/(2\sigma^2)\}$
with $\sigma=1$
for all $\mathcal{H}_k$
due to 
the unit-variance constraint
in \eqref{orignal multiview obj} (see Section~\ref{sec: RKHS}),
and
reparametrize 
$\bd{u}_k$
as
$\mb{\Pi}_k=\bd{u}_k \bd{u}_k^\top$
for the optimization problem, leading to the following optimization formulation:
\begin{align}
     &   \max_{\{\mb{\Pi}_k\in \mathcal{M}_k \}_{k=1}^K}     
     \sum_{1\le s<t\le K}      
     H(\mb{\Pi}_s,\mb{\Pi}_t) - \sum_{k=1}^K{\lambda_k \| \mb{\Pi}_k\|_1}
     \label{multiview obj}\\
  &      \st \quad
 \tr(\mb{\Sigma}_k^{1/2}\mb{\Pi}_k   \mb{\Sigma}_k^{1/2})=1,
 \quad
\text{rank}(\mb{\Pi}_k)=1,
  \nonumber
\end{align}
where $\mathcal{M}_k$ is the set of $p_k\times p_k$  symmetric positive semi-definite real-valued matrices, 
\begin{align*}
H(\mb{\Pi}_s,&\mb{\Pi}_t)
:=\mathrm{HSIC}(\bd{u}_s^\top\bd{x}_s,\bd{u}_t^\top\bd{x}_t;\mathcal{H}_s,\mathcal{H}_t) \\
&=  \text{E}[\exp(-\frac{1}{2}\langle \mb{\Pi}_s, \mb{Z}_s\rangle)
\exp(-\frac{1}{2}\langle \mb{\Pi}_t, \mb{Z}_t\rangle)]\\
&\quad +\text{E}[\exp(-\frac{1}{2}\langle \mb{\Pi}_s, \mb{Z}_s\rangle)]
\text{E}[\exp(-\frac{1}{2}\langle \mb{\Pi}_t, \mb{Z}_t\rangle)]
\\
&\quad -2\mathrm{E}\big[\mathrm{E}[\exp(-\frac{1}{2}\langle \mb{\Pi}_s, \mb{Z}_s\rangle)|\bd{x}_s]\mathrm{E}[
\exp(-\frac{1}{2}\langle \mb{\Pi}_t, \mb{Z}_t\rangle)|\bd{x}_t]\big]
\end{align*}
due to \eqref{eqn: HSIC in kernel},
$\langle \mb{\Pi}_k, \mb{Z}_k\rangle=\tr(\mb{\Pi}_k^\top\mb{Z}_k)$,
$\mb{Z}_k=(\bd{x}_k-\bd{x}_k')(\bd{x}_k-\bd{x}_k')^\top$,
and 
$\{\bd{x}_k'\}_{k=1}^K$ is an i.i.d. copy of $\{\bd{x}_k\}_{k=1}^K$.
The same reparameterization 
is used in 
sparse principal component analysis
\citep{MR3546438}, 
sparse sliced inverse regression \citep{tan2018convex},
 and sparse single-index regression \citep{chen2024high}.
 
However,
the constraint sets in \eqref{multiview obj}
are not convex due to the equality on
the rank function.
By noting
$
\text{rank}(\mb{\Sigma}_k^{1/2}\mb{\Pi}_k\mb{\Sigma}_k^{1/2})
=
\text{rank}(\mb{\Pi}_k)=1
$,
we instead use
their  Fantope-like
convex relaxations
$
\{\mb{\Pi}_k\in \mathcal{M}_k:
\tr(\mb{\Sigma}_k^{1/2}\mb{\Pi}_k   \mb{\Sigma}_k^{1/2})=1
\}$, $k=1,\dots, K$ \citep{vu2013fantope,overton1992sum}.
Thus, we first solve
the relaxed optimization:
\begin{align}
   \{\mb{\Pi}_k^*\}_{k=1}^K = &   \underset{\{\mb{\Pi}_k\in \mathcal{M}_k \}_{k=1}^K}{\argmax} 
     \sum_{1\le s<t\le K}      
     H(\mb{\Pi}_s,\mb{\Pi}_t) - \sum_{k=1}^K{\lambda_k \| \mb{\Pi}_k\|_1}
   \nonumber\\
  &      \st \quad
 \tr(\mb{\Sigma}_k^{1/2}\mb{\Pi}_k   \mb{\Sigma}_k^{1/2})=1,
  \label{relaxed multiview obj}
\end{align}
and then 
use the top eigenvector $\bd{u}_k^*$ of 
$\mb{\Pi}_k^*$, scaled to satisfy $(\bd{u}_k^*)^\top\mb{\Sigma}_k\bd{u}_k^*=1$, to approximate the optimal $\bd{u}_k$
in 
problem~\eqref{orignal multiview obj}.

Empirically, 
with 
 $n$ i.i.d. observations 
$\{\bd{x}_1^{(i)},\dots,\bd{x}_K^{(i)}\}_{i=1}^{n}$
of $\{\bd{x}_1,\dots, \bd{x}_K\}$,
substituting the empirical HSIC from \eqref{eqn: empirical HSIC} and
a covariance matrix estimator
$\widehat{\mb{\Sigma}}_k$
for the population HSIC and the covariance matrix $\mb{\Sigma}_k$
in~\eqref{relaxed multiview obj}
yields
the estimators of $\{\mb{\Pi}_k^*\}_{k=1}^K$:
\begin{align}\label{est relaxed multiview obj}
\{ \widehat{\mb{\Pi}}_k\}_{k=1}^K
=
&\underset{\{\mb{\Pi}_k\in \mathcal{M}_k \}_{k=1}^K}{\argmin} 
 -\sum_{1\le s<t\le K}
 \frac{\tr\big(\mb{K}_s(\mb{\Pi}_s)\mb{H}\mb{K}_t(\mb{\Pi}_t)\mb{H}\big)}{n^2}
\nonumber\\
& 
~ + \sum_{k=1}^K{\lambda_k \| \mb{\Pi}_k\|_1}
 ~~ \st ~~
 \tr(\widehat{\mb{\Sigma}}_k^{1/2}\mb{\Pi}_k   \widehat{\mb{\Sigma}}_k^{1/2})=1.
\end{align}
Here, 
$\mb{K}_k(\mb{\Pi}_k)
\in \mathbb{R}^{n\times n}
$
has 
$(i,j)$-th entry
$\mb{K}_k^{[i,j]}(\mb{\Pi}_k)
=\exp(-\langle \mb{\Pi}_k,\mb{Z}_k^{(ij)} \rangle/2)$,
with 
$\mb{Z}_k^{(ij)}=(\bd{x}_k^{(i)}-\bd{x}_k^{(j)})(\bd{x}_k^{(i)}-\bd{x}_k^{(j)})^\top$.
We define 
$\widehat{\mb{\Sigma}}_k=(1-\epsilon_k)\mb{S}_k+\epsilon_k \mb{I}_{p_k}$,
where $\mb{S}_k$ 
is the sample covariance matrix of $\bd{x}_k$, 
and $\epsilon_k\ge 0$ is a very small constant
such that $\widehat{\mb{\Sigma}}_k$
is  invertible and
very close to $\mb{S}_k$ when $\mb{S}_k$ is singular \citep{ledoit2004honey}. 
Notably, $\mb{S}_k$ 
is singular 
if $p_k>n$.
We set 
$
\epsilon_k=10^{-4}\|\mb{S}_k\|_F/\|\mb{I}_{p_k}-\mb{S}_k\|_F
$
if $\mb{S}_k$ is singular;
otherwise, $\epsilon_k=0$.
The invertibility of $\widehat{\mb{\Sigma}}_k$
ensures
 the equivalence of 
 $\mb{\Pi}_k\in \mathcal{M}_k$
and
$\widehat{\mb{\Sigma}}_k^{1/2}\mb{\Pi}_k   \widehat{\mb{\Sigma}}_k^{1/2}\in \mathcal{M}_k$, 
 facilitating our algorithm development.
Finally, we use
the top eigenvector $\widehat{\bd{u}}_k$ of 
$\widehat{\mb{\Pi}}_k$, scaled to satisfy $\widehat{\bd{u}}_k^\top\mb{S}_k\widehat{\bd{u}}_k=1$, as the estimator for $\bd{u}_k^*$.

\subsection{HSIC-SGCCA Algorithm Development}
We propose an efficient algorithm 
for solving the optimization problem~\eqref{est relaxed multiview obj}, which remains nonconvex 
and is not even multi-convex.
We apply  BPL \citep{xu2017globally} to solve this nonconvex problem, with each subproblem within BPL optimized via LADMM \citep{fang2015generalized}.
{\color{black} Unlike BCD,
BPL alternately updates each block of variables
by minimizing a prox-linear surrogate function
instead of the original objective,
eliminating the need for any convexity assumptions such as multi-convexity. }

Let $f(\{\mb{\Pi}_k\}_{k=1}^K)=
 -  \sum_{1\le s<t\le K}
\frac{1}{n^2}\tr\big(\mb{K}_s(\mb{\Pi}_s)\mb{H}\mb{K}_t(\mb{\Pi}_t)\mb{H}\big)$.
The $(r+1)$-th iteration 
of BPL 
updates $\mb{\Pi}_k$
for all $k=1,\dots, K$~by 
\begin{align}\label{eqn: BLP}
\mb{\Pi}_k^{(r+1)}=
&\argmin_{\mb{\Pi}_k\in \mathcal{D}_k}
\langle  \mb{\Pi}_k, \nabla f_k^{(r)}(\mb{\Pi}_k^{(r)}) \rangle \\
&\quad +\frac{L_k^{(r)}}{2} \|\mb{\Pi}_k-\mb{\Pi}_k^{(r)} \|_F^2 
+\lambda_k\|\mb{\Pi}_k \|_1,
\nonumber
\end{align}
where 
$\mathcal{D}_k=\{\mb{\Pi}_k\in \mathcal{M}_k: \tr(\widehat{\mb{\Sigma}}_k^{1/2}\mb{\Pi}_k   \widehat{\mb{\Sigma}}_k^{1/2})=1 \}$,
\begin{align*}
\nabla f_k^{(r)}(\mb{\Pi}_k)&:=
\nabla_{\mb{\Pi}_k} f( 
\{\mb{\Pi}_\ell^{(r+1)}\}_{\ell<k}
, \mb{\Pi}_k,\{\mb{\Pi}_\ell^{(r)}\}_{\ell> k})
\nonumber\\
&=\frac{1}{2n^2}
\sum_{i,j=1}^n \exp\left(
-\frac{\langle \mb{\Pi}_k, \mb{Z}_k^{(ij)}  \rangle}{2}
\right) (\widetilde{\mb{K}}_{-k}^{(r)})^{[i,j]}    
\mb{Z}_k^{(ij)},
\end{align*}  
$(\widetilde{\mb{K}}_{-k}^{(r)})^{[i,j]}$
is the $(i,j)$ entry
of the matrix 
\be
\widetilde{\mb{K}}_{-k}^{(r)}
:=\mb{H}
[\sum_{\ell< k} \mb{K}_\ell(\mb{\Pi}_\ell^{(r+1)})  
+\sum_{\ell> k} \mb{K}_\ell(\mb{\Pi}_\ell^{(r)}) 
]
\mb{H},
\ee
and 
$L_k^{(r)}$ is a BPL parameter larger than the Lipschitz constant, $\widetilde{L}_k^{(r)}$, of 
$\nabla f_k^{(r)}(\mb{\Pi}_k)$,
 defined such that
$
\|\nabla f_k^{(r)}(\mb{\Pi}_k)-\nabla f_k^{(r)}(\widetilde{\mb{\Pi}}_k)\|_F
\le 
\widetilde{L}_k^{(r)}\|\mb{\Pi}_k- \widetilde{\mb{\Pi}}_k\|_F  
$
for any $\mb{\Pi}_k,\widetilde{\mb{\Pi}}_k\in \mathcal{D}_k$.
We~have
\be\label{Lconstant}
\widetilde{L}_k^{(r)}=\frac{1}{4n^2}\sum_{i,j=1}^n 
|(\widetilde{\mb{K}}_{-k}^{(r)})^{[i,j]}|\| \mb{Z}_k^{(ij)}\|_F^2.
\ee
The sequence $\{\mb{\Pi}_1^{(r)},\dots, 
\mb{\Pi}_K^{(r)}\}_{r\ge 1}$
generated from \eqref {eqn: BLP}
by the BLP method 
ensures a monotone decrease of the objective function
in~\eqref{est relaxed multiview obj}
and converges to a critical point \citep{xu2017globally}.

To solve the subproblem~\eqref{eqn: BLP},
we equivalently write it as
\be\label{eqn: BLP quadratic}
\min_{\mb{\Pi}_k\in\mathcal{D}_k}\frac{L_k^{(r)}}{2}\Big\| \mb{\Pi}_k-\Big[\mb{\Pi}_k^{(r)}-\frac{1}{L_k^{(r)}}\nabla f_k^{(r)}(\mb{\Pi}_k^{(r)})\Big]\Big\|_F^2
+\lambda_k\|\mb{\Pi}_k \|_1.
\ee
This subproblem is a convex, penalized quadratic problem, which ensures that any local minimum is also a global minimum. 
The difficulty in directly solving 
\eqref{eqn: BLP quadratic} is the interaction between 
the penalty term and the constraint 
$\mathcal{D}_k$.
We apply the LADMM \citep{fang2015generalized,chen2024high}
to solve it.
We first rewrite it 
as 
\begin{align}\label{eqn: BLP separate}
\min_{\mb{\Pi}_k,\mb{H}_k}
&\frac{L_k^{(r)}}{2}\Big\| \mb{\Pi}_k-\Big[\mb{\Pi}_k^{(r)}-\frac{1}{L_k^{(r)}}\nabla f_k^{(r)}(\mb{\Pi}_k^{(r)})\Big]\Big\|_F^2
+\lambda_k\|\mb{\Pi}_k \|_1\nonumber\\
&~+\infty \cdot \mathbb{I}(\mb{H}_k\not\in \widetilde{\mathcal{D}}_k)
\quad \st \quad \widehat{\mb{\Sigma}}_k^{1/2}\mb{\Pi}_k   \widehat{\mb{\Sigma}}_k^{1/2}=\mb{H}_k,
\end{align}
where $\widetilde{\mathcal{D}}_k=\{\mb{H}_k\in \mathcal{M}_k:\tr(\mb{H}_k)=1\}$,
$\mathbb{I}(\cdot)$ is the indicator function,
and we define $\infty\cdot 0=0$. 
Since $\widehat{\mb{\Sigma}}_k$ is invertiable due to the use of $\epsilon_k$, 
$\mb{H}_k\in \widetilde{\mathcal{D}}_k$
is equivalent to $\mb{\Pi}_k\in\mathcal{D}_k$.
The scaled augmented Lagrangian (AL) function
for~\eqref{eqn: BLP separate}~is 
\begin{align*}
&\mathcal{L}(\mb{\Pi}_k,\mb{H}_k;\mb{\Gamma}_k,\rho_k)=\\
&
\frac{L_k^{(r)}}{2}\Big\| \mb{\Pi}_k-\Big[\mb{\Pi}_k^{(r)}-\frac{1}{L_k^{(r)}}\nabla f_k^{(r)}(\mb{\Pi}_k^{(r)})\Big]\Big\|_F^2
+\lambda_k\|\mb{\Pi}_k \|_1\\
&+\infty \cdot \mathbb{I}(\mb{H}_k\not\in \widetilde{\mathcal{D}}_k)
+\frac{\rho_k}{2} (\|\widehat{\mb{\Sigma}}_k^{1/2}\mb{\Pi}_k   \widehat{\mb{\Sigma}}_k^{1/2}-\mb{H}_k+\mb{\Gamma}_k\|_F^2-\|\mb{\Gamma}_k \|_F^2),
\end{align*}
where 
$\mb{\Gamma}_k$ is the scaled dual variable, 
and $\rho_k>0$ is the AL parameter.
The LADMM  minimizes
$\mathcal{L}(\mb{\Pi}_k,\mb{H}_k;\mb{\Gamma}_k,\rho_k)$
 by alternatively
updating
$\{\mb{\Pi}_k,\mb{H}_k,\mb{\Gamma}_k\}$
with closed-form updates:
\begin{align}
\mb{\Pi}_k^{(r,j+1)}
&=\text{Soft}\Big(
\frac{\tau_k}{L_k^{(r)}+\tau_k}
\Big[
\mb{\Pi}_k^{(r,j)}
-\frac{\rho_k}{\tau_k}
\widehat{\mb{\Sigma}}_k
\mb{\Pi}_k^{(r,j)}
\widehat{\mb{\Sigma}}_k
\label{LADMM: Pi_k}\\
&\qquad\qquad +\frac{\rho_k}{\tau_k}
\widehat{\mb{\Sigma}}_k^{1/2}
(\mb{H}_k^{(r,j)}-\mb{\Gamma}_k^{(r,j)})\widehat{\mb{\Sigma}}_k^{1/2}
\Big]\nonumber\\
&\qquad\qquad+
\frac{L_k^{(r)}}{L_k^{(r)}+\tau_k}
\Big[\mb{\Pi}_k^{(r)}-\frac{1}{L_k}\nabla f_k^{(r)}(\mb{\Pi}_k^{(r)})\Big],\nonumber\\
&\qquad\qquad\frac{\lambda_k}{L_k^{(r)}+\tau_k}
\Big),
\nonumber\\
\mb{H}_k^{(r,j+1)}
&=\mathcal{P}_{\widetilde{\mathcal{D}}_k}
(\widehat{\mb{\Sigma}}_k^{1/2}\mb{\Pi}_k^{(r,j+1)}\widehat{\mb{\Sigma}}_k^{1/2}+\mb{\Gamma}_k^{(r,j)}),\label{LADMM: H_k}\\
\mb{\Gamma}_k^{(r,j+1)}
&=\mb{\Gamma}_k^{(r,j)}
+\widehat{\mb{\Sigma}}_k^{1/2}\mb{\Pi}_k^{(r,j+1)}\widehat{\mb{\Sigma}}_k^{1/2}-\mb{H}_k^{(r,j+1)},
\label{LADMM: Gamma_k}
\end{align}
where $\text{Soft}$ is the entrywise soft thresholding 
such that
the \mbox{$(i,j)$-th} entry of
$\text{Soft}(\mb{M},T)$
is
$\text{sign}(M_{ij})\cdot\max(|M_{ij}|-T,0)$
for any matrix $\mb{M}=(M_{ij})$ and threshold $T>0$,
$\tau_k>0$ is the LADMM parameter,
and
$\mathcal{P}_{\widetilde{\mathcal{D}}_k}$
is the Euclidean projection onto $\widetilde{\mathcal{D}}_k$.
For any symmetric matrix $\mb{W}\in \mathbb{R}^{p_k\times p_k}$,
$
\mathcal{P}_{\widetilde{\mathcal{D}}_k}
(\mb{W})
=\sum_{i=1}^{p_k}
w_i^+\bd{v}_i\bd{v}_i^\top
$,
where $w_i^+=\max(w_i-\theta,0)$,
$\theta=\frac{1}{m}(\sum_{i=1}^m w_i-1)$,
$m=\max\{j: w_j-\frac{1}{j}(\sum_{i=1}^j w_i-1)>0\}$,
and
$\mb{W}=\sum_{i=1}^{p_k}w_i \bd{v}_i\bd{v}_i^\top$ is the eigen-decomposition of $\mb{W}$ with eigenvalues $w_1\ge \dots \ge w_{p_k}$
\citep{vu2013fantope,duchi2008efficient}.

Algorithm~\ref{algorithm table}
summarizes  the HSIC-SGCCA algorithm.
In our numerical studies,
we use
the BLP parameter
$L_k^{(r)}=2\widetilde{L}_k^{(r)}$
with $\widetilde{L}_k^{(r)}$  in
\eqref{Lconstant} 
\citep{xu2017globally},
and set
the AL parameter
$\rho_k=1$ and 
the LADMM parameter $\tau_k=4\rho_k \| \widehat{\mb{\Sigma}}_k\|_2^2$  \citep{fang2015generalized,chen2024high}.
{\color{black}The computational complexity of
the algorithm without tuning is
$O(R(J\sum_{k=1}^Kp_k^3+
n^2\sum_{k=1}^Kp_k^2))$}.
We use five-fold cross-validation to tune the sparsity parameters $\{\lambda_k\}_{k=1}^K$
and adopt the routine multi-start strategy \citep{mart2018handbook} to mitigate the issue of BPL convergence to a critical point that is not a global optimum;
see details in the Supplementary Material.

\begin{algorithm}[!t]
\caption{HSIC-SGCCA algorithm}\label{algorithm table}
\begin{algorithmic}[1]
\INPUT Multi-view data $\{\mb{X}_k\}_{k=1}^K$, tuning parameters $\{\lambda_k\}_{k=1}^K$, convergence tolerance $\epsilon>0$,
maximum numbers of iterations $R$ and $J$.
\State Initialize $\mb{\Pi}_k^{(0)}\in \mathcal{D}_k,k=1\dots, K$;
\Repeat $\quad r= 0,1,2,\dots$
\For{$k=1,\dots, K$}
\State $\mb{\Pi}_k^{(r,0)}=\mb{\Pi}_k^{(r)},
\mb{H}_k^{(r,0)}=
\widehat{\mb{\Sigma}}_k^{1/2}\mb{\Pi}_k^{(r,0)}   \widehat{\mb{\Sigma}}_k^{1/2}$, 
$\mb{\Gamma}_k^{(r,0)}=\mb{0}$;
\Repeat $\quad j = 0,1,2,\dots$
    \State Update $(\mb{\Pi}_k^{(r,j+1)},\mb{H}_k^{(r,j+1)},
    \mb{\Gamma}_k^{(r,j+1)})$
    by \eqref{LADMM: Pi_k}-\eqref{LADMM: Gamma_k};
\Until 
$\max\{\| \mb{\Pi}_k^{(r,j+1)}-\mb{\Pi}_k^{(r,j)} \|_{\max},$

\qquad\qquad\qquad $\|\widehat{\mb{\Sigma}}_k^{1/2}\mb{\Pi}_k^{(r,j+1)}\widehat{\mb{\Sigma}}_k^{1/2}-\mb{H}_k^{(r,j+1)} \|_{\max}
\}\le \epsilon
$

~~~~~\qquad
or $j\ge J$;
\State Update $\mb{\Pi}_k^{(r+1)}=\mb{\Pi}_k^{(r,j+1)}$;
\EndFor
\Until 
$\max_{1\le k\le K}\| \mb{\Pi}_k^{(r+1)}-\mb{\Pi}_k^{(r)} \|_{\max}
\le \epsilon$
or $r\ge R$;
\OUTPUT $\widehat{\bd{u}}_k=$
the top eigenvector of $\mb{\Pi}_k^{(r+1)}$, scaled to satisfy $\widehat{\bd{u}}_k^\top\mb{S}_k\widehat{\bd{u}}_k=1$, for $k=1,\dots, K$.
\end{algorithmic}
\end{algorithm}

\section{Simulations}\label{sec: simulation}
We conduct simulations to compare the variable selection performance of 
our  HSIC-SGCCA against
 nonlinear GCCA methods including
our  SA-KGCCA and TS-KGCCA, and DGCCA \citep{benton2017deep},
as well as linear SGCCA methods
including
$\ell_1$-penalized SUMCOR-SGCCA \citep{kanatsoulis2018structured} and 
$\ell_1$-minimized MAXVAR-SGCCA \citep{lv2024sparse}.
Since DGCCA does not impose sparsity, we perform its variable selection by identifying variables with absolute  importance scores above 0.05. 
Each variable's importance score is first computed as the change in the loss function
of the trained DGCCA model when the variable is set to its sample mean \citep{NEURIPS2020_41c542df}, and is  then scaled  so that the 
$\ell_2$ norm of the importance score vector within each data view equals one.
The implementation details
 of
 these methods
 are provided in the Supplementary Material.

\vspace{-0.15cm}
\subsection{Simulation settings}
We consider three-view data
$\mb{X}_k\in\mathbb{R}^{p\times n}, k\in\{1,2,3\}$,
where 
the $n$ columns of $\mb{X}_k$ are i.i.d.
copies 
of the random vector $\bd{x}_k$
generated
from the following two models. 
Similar models are considered in \citet{tenenhaus2014variable,tenenhaus2015kernel}.

\begin{itemize}
\item {\bf Linear model:} $\bd{x}_{k}^{[j]}=v_k \mathbb{I}( j\le q)+\epsilon_{kj}$ for $j\in\{1,\dots,p\}$ and $k\in\{1,2,3\}$,
where $(v_1,v_2,v_3)^\top$
follows a trivariate 
Gaussian distribution with 
zero mean,
$\var(v_k)=1$, 
$\cov(v_1,v_3)=\cov(v_2,v_3)=0.7$
and $\cov(v_1,v_2)=0$,
and $\{\epsilon_{kj}\}_{k,j}$ 
are i.i.d. Gaussian variables with zero mean
and variance 0.2.

\item {\bf Nonlinear model:} 
$\bd{x}_1^{[j]}=v\mathbb{I}( j\le q)+\epsilon_{1j}$,
$\bd{x}_2^{[j]}=v^2\mathbb{I}( j\le q)+\epsilon_{2j}$,
$\bd{x}_3^{[j]}=v\cos(v)\mathbb{I}( j\le q)+\epsilon_{3j}$
for 
$j\in\{1,\dots, p\}$,
where $v$ follows a uniform distribution
on $[-2\pi, 2\pi]$,
and $\{\epsilon_{kj}\}_{k,j}$ 
are i.i.d. Gaussian variables with zero mean
and variance 0.2. 

\end{itemize}

We consider 
the above two models
under the settings
with (i) varying 
$p\in \{30,50,100,200\}$
and fixed $(n,q)=(100,5)$,
(ii) varying $n\in \{100,200,400\}$
and fixed $(p,q)=(100,5)$,
and (iii) varying 
$q\in\{5,10,20\}$
 and fixed $(p,n)=(100,100)$.
Note that the impact of the total variable dimension ($3p$) of the three data views, not just the variable dimension ($p$) of each single data view, should be considered in variable selection, as all variables contribute to the estimations in GCCA methods \citep{laha2022support}.
Each simulation setting is
conducted for 100 independent replications.

\subsection{Evaluation metrics}
We evaluate the variable selection performance of the six GCCA methods using six classification metrics \citep{chicco2020advantages}:  
F1 score, Matthews correlation coefficient (MCC), precision, recall (i.e., sensitivity), specificity,
and success rate (SR).
For each data view,
since the first $q$ variables
contain a shared component
and the six methods
use different selection criteria,
we create a single joint label
for these $q$ variables:
the true label value is positive;
the predicted label is positive if at least one of the $q$ variables is selected, and otherwise is negative.
Each of the remaining
$p-q$ variables in each data view
has a label with a true value of negative,
and its predicted value is positive if selected.
The six classification metrics are computed
based on the pooled result from the three data views.
The success rate is computed over the
100 simulation replications.
The variable selection in a simulation replication is considered successful if, for each data view,
at least one of the first $q$ variables
is selected and none of the remaining $p-q$ variables are selected. 
We also evaluate the timing performance by  measuring the runtime of each algorithm with the optimal tuning parameters applied.

\begin{figure*}[t!]
\centering
    \begin{subfigure}{0.33\linewidth}
    \includegraphics[width=\linewidth]{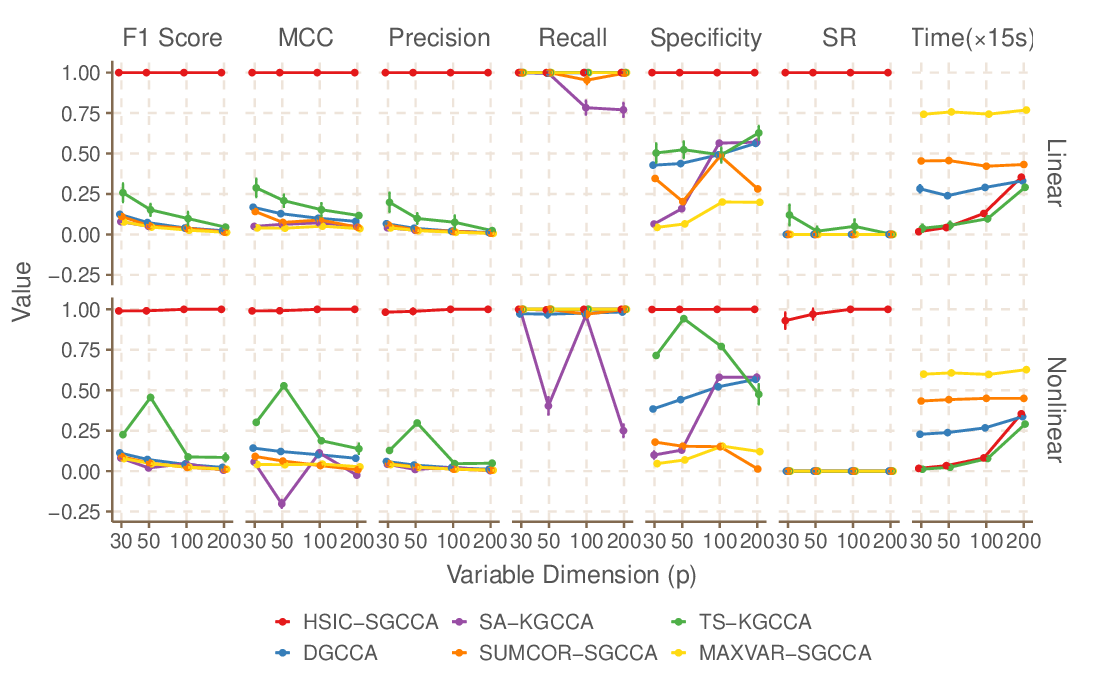}
    \caption{$p\in \{30,50,100,200\}$ and $(n,q)=(100,5)$.}
    \label{fig1a}
    \end{subfigure}
    \begin{subfigure}{0.33\linewidth}
    \includegraphics[width=\linewidth]{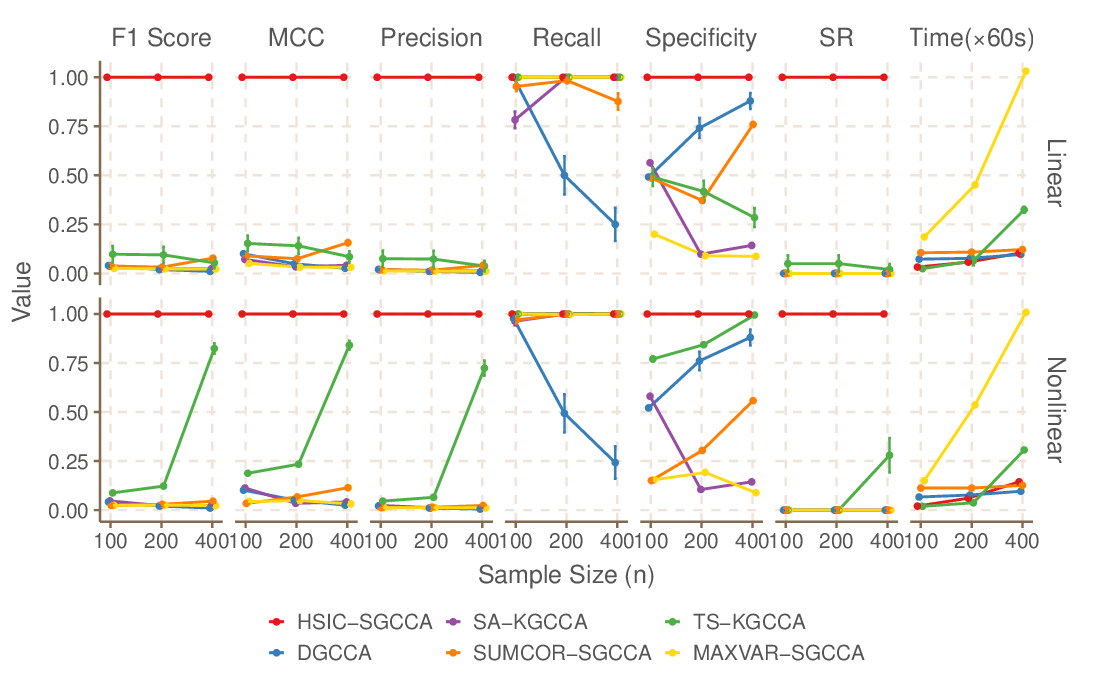}
    \caption{$n\in \{100,200,400\}$ and 
$(p,q)=(100,5)$.}
    \label{fig1b}
    \end{subfigure}
    \begin{subfigure}{0.33\linewidth}
    \includegraphics[width=\linewidth]{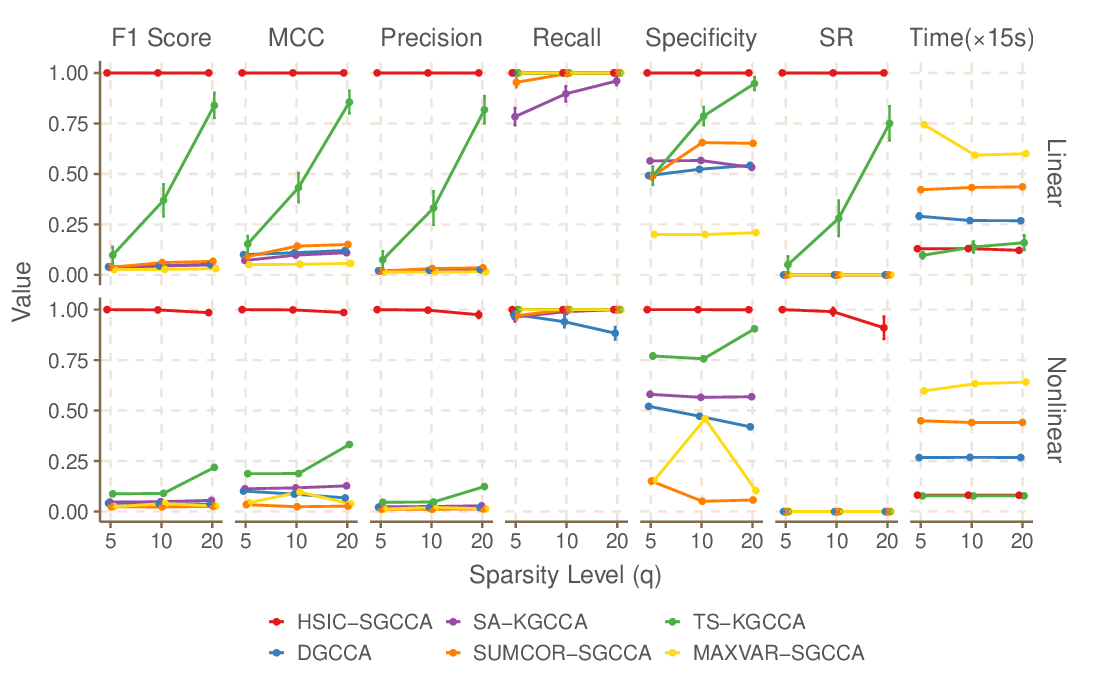}
    \caption{$q\in\{5,10,20\}$
 and $(p,n)=(100,100)$.}
    \label{fig1c}
    \end{subfigure}
    
    \vspace{-0.3cm}
\caption{
Simulation results (Average $\pm$ 1.96SE)
based on 100 independent replications.
Small error bars may be hidden by the average dots. 
Runtime results of SA-KGCCA are provided in 
Supplementary Table~S1 due to large values.
}\label{fig2}
\end{figure*}

\subsection{Simulation results}

Figure~\ref{fig2} 
summarizes the variable selection performance of the six  methods
in the seven metrics. Results are presented as
the average $\pm$ 1.96 standard error (SE) of the mean over the 100 simulation replications. 
The runtime of SA-KGCCA is provided in 
Supplementary Table S1 due to substantially larger values than other methods.
Overall, our HSIC-SGCCA demonstrates the best performance in both linear and nonlinear models,
achieving nearly perfect variable selection in all settings with a significant margin while maintaining a speed comparable to the fastest methods.

Specifically, Figure~\ref{fig1a}
shows the results 
for varying 
$p\in \{30,50,100,200\}$
with  $(n,q)=(100,5)$.
In the linear model,
our HSIC-SGCCA  achieves perfect variable selection performance.
The recall values are close to one for all methods
except SA-KGCCA, indicating that
most positives are correctly identified.
However, the low precision values for all  methods
except HSIC-SGCCA show that they
predict more false positives than true positives. 
As the variable dimension 
$p$ of each data view increases,  precision declines, even though specificity  improves (for all except SUMCOR-SGCCA) due to the data imbalance, with  relatively few fixed positives and a growing number of negatives.
 F1 score and MCC,
 which provide a more comprehensive evaluation in imbalanced data settings, remain below
  0.55 for all five methods other than HSIC-SGCCA.
The SR, a stricter metric,
shows near-zero values for these five methods.
Surprisingly,  the linear SGCCA methods,
SUMCOR-SGCCA and MAXVAR-SGCCA,
fail to perform well in the linear model.
In terms of runtime, HSIC-SGCCA is relatively fast,
significantly outperforming 
the two linear methods
and is comparable to the
fastest method, TS-KGCCA, which is based on soft-thresholding.

For the more challenging nonlinear model shown in Figure~\ref{fig1a},
our HSIC-SGCCA continues to demonstrate the best performance. In contrast, 
the other three nonlinear methods  (SA-KGCCA, TS-KGCCA, and DGCCA)
and the two linear models 
(SUMCOR-SGCCA and MAXVAR-SGCCA)
still perform poorly.
 The runtime of all methods remains similar to that in the linear model.


Figure~\ref{fig1b} presents
 the results 
for varying 
$n\in \{100,200,400\}$ with 
$(p,q)=(100,5)$.
HSIC-SGCCA consistently achieves  perfect variable selection.
As the sample size $n$ increases to 400,
SUMCOR-SGCCA shows notable improvement in the linear model, while TS-KGCCA exhibits significant gains in the nonlinear model.
In contrast,
the two methods in the other settings and the remaining methods yield poor results.
Figure~\ref{fig1c} illustrates
the results for varying $q\in\{5,10,20\}$
 with $(p,n)=(100,100)$.
As the sparsity level $q$ increases, none of the methods  experience   a 
substantial decline in  performance
in terms of F1 score and MCC. 

\vspace{-0.2cm}
\section{Applications to TCGA-BRCA data}\label{sec: real data}

We apply the six aforementioned  GCCA methods to
 breast invasive carcinoma data from TCGA  \citep{Kobo12}.
We use the
three-view data from 
a common set of 
$n=1057$ primary solid tumor samples
from 1057 female patients, including
mRNA expression data 
for  the $p_1= 2596$ most
variably expressed genes,
DNA methylation data for  the $p_2=3077$ most variably methylated CpG sites,
and 
miRNA expression data for  the $p_3=523$ most variably expressed miRNAs.
The tumor samples are categorized  into five PAM50 intrinsic subtypes  \citep{parker2009supervised},
including 182 Basal-like,  552 Luminal A, 202 Luminal B,
81 HER2-enriched, and 40 normal-like tumors.
All variables are standardized to have zero mean and unit variance. 
DGCCA is applied with
the same variable selection approach 
used in the simulations. 
The data preprocessing and implementation details 
of our analysis are provided in the Supplementary Material.


\begin{table}[!t]
\caption{Counts of selected variables in TCGA-BRCA data.\label{tabvs}}%
\begin{tabular*}{\columnwidth}{@{\extracolsep\fill}lccc@{\extracolsep\fill}}
\hline
& mRNA  & DNA  & miRNA \\
&expression & methylation & expression\\
Method & (genes) & (CpG sites) & (miRNAs) \\
\hline
All variables         & 2596  & 3077 & 523\\
 HSIC-SGCCA           & 15 & 24 & 32 \\ 
 SA-KGCCA           & 213 & 1 & 1 \\ 
TS-KGCCA            & 217 & 221 & 461 \\
DGCCA            & 87 & 80 & 82\\ 
 SUMCOR-SGCCA     & 2558 & 3043 & 69 \\
 MAXVAR-SGCCA     & 19 & 16 & 102 \\
\hline
\end{tabular*}\vspace{-0.4cm}
\end{table}

\begin{figure*}[b!]
\centering
    \begin{subfigure}{0.33\linewidth}
        \centering
        \includegraphics[width=\linewidth]{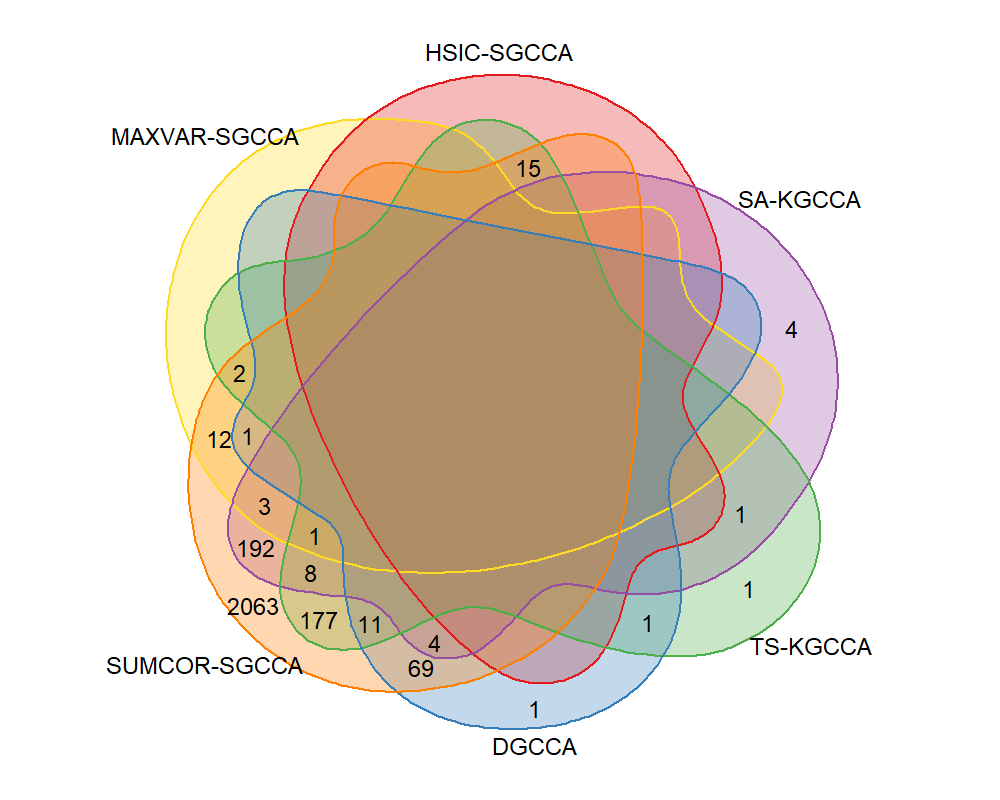}
        \caption{mRNA expression (genes)}
        \label{figvenna}
    \end{subfigure}
    \hfill
    \begin{subfigure}{0.33\linewidth}
        \centering
        \includegraphics[width=\linewidth]{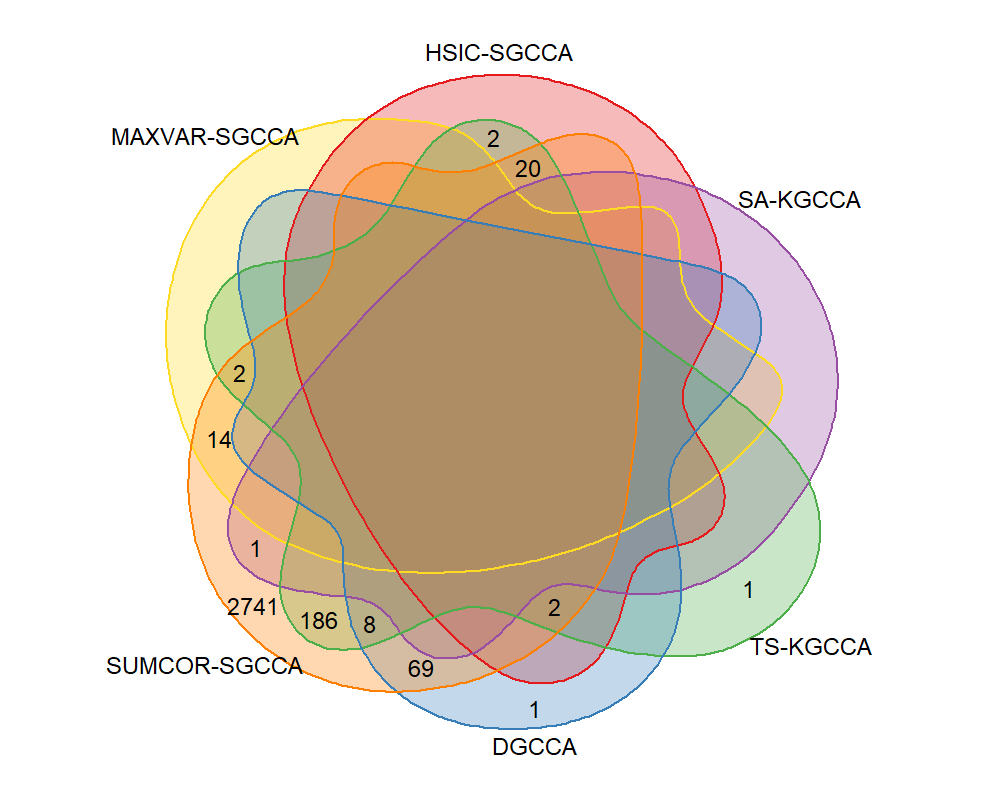}
        \caption{DNA methylation (CpG sites)}
        \label{figvennb}
    \end{subfigure}
    \hfill
    \begin{subfigure}{0.33\linewidth}
        \centering
        \includegraphics[width=\linewidth]{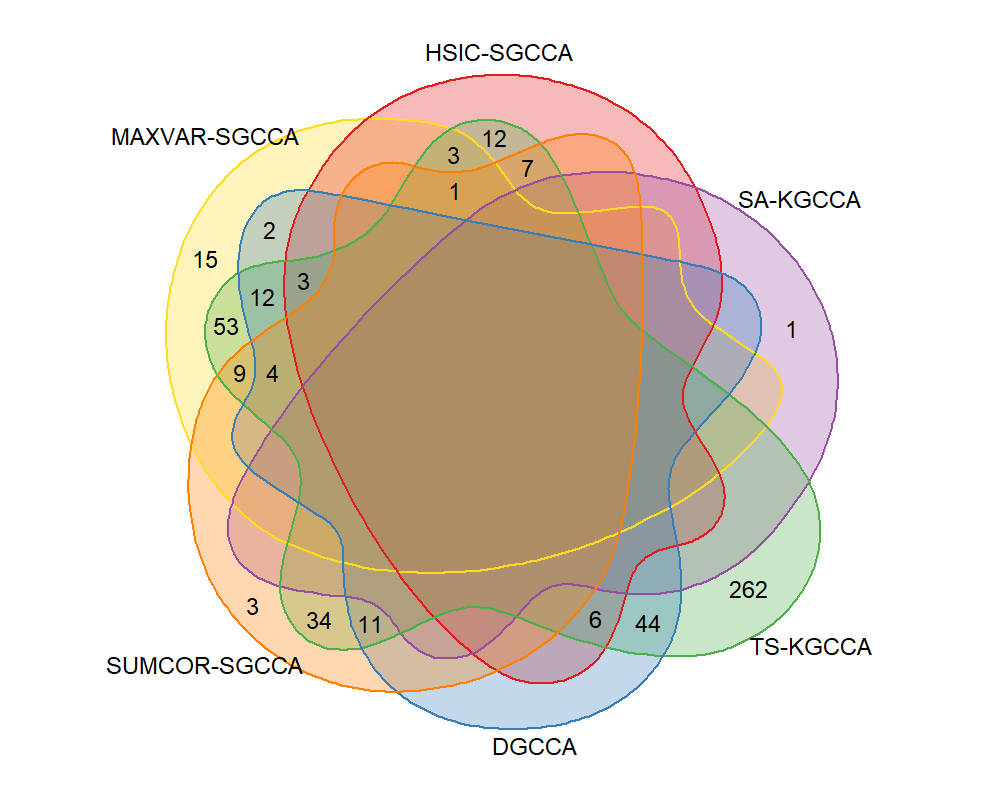}
        \caption{miRNA expression (miRNAs)}
        \label{figvennc}
    \end{subfigure}

    \vspace{-0.4cm}
\caption{Venn diagrams showing the numbers of TCGA-BRCA variables selected by the six methods. Total counts are provided in Table~\ref{tabvs}.}
\label{figvenn}
\end{figure*}

Table \ref{tabvs} shows the counts
of  variables selected  by each method. Our HSIC-SGCCA
identifies 
a subset of reasonable size, including 
15 genes, 24 CpG sites, and 32 miRNAs. 
DGCCA  selects approximately 80 variables in each data view, whereas MAXVAR-SGCCA finds 
19 genes, 16 CpG sites,
and 102 miRNAs.
In contrast,
SUMCOR-SGCCA retains nearly all genes and CpG sites along with 69 miRNAs, while
SA-KGCCA  selects only
1 CpG site and 1 miRNA but identifies 213 genes. TS-KGCCA eliminates over 90\% of both genes and CpG sites but retains 88\% of the miRNAs.
Figure \ref{figvenn}  presents  Venn diagrams  illustrating  the overlaps among these selections.
All variables selected by HSIC-SGCCA 
are also chosen by at least one other method, even when excluding SUMCOR-SGCCA, which selects nearly all genes and CpG sites. Among
the 32 miRNAs selected by HSIC-SGCCA,
12 are shared only with
 TS-KGCCA, which retains 88\% of the miRNAs.
In contrast, 69 of the 87 genes and 69 of the 80 CpG sites selected by DGCCA are also identified only by SUMCOR-SGCCA, and 44 of the 82 DGCCA-selected miRNAs overlap exclusively with TS-KGCCA.
DGCCA also selects
1 gene and
1 CpG site  that no other method identifies.
For MAXVAR-SGCCA, 12 of its 19 selected genes and 14 of its 16 CpG sites are chosen solely by SUMCOR-SGCCA,  53 of its 102 miRNAs are shared only with TS-KGCCA, and 15 miRNAs are not selected by any other method.
Overall, HSIC-SGCCA achieves a balanced performance in variable selection by identifying a reasonable number of variables with meaningful overlaps with other methods.

\begin{table*}[t!]
\caption{Performance in PAM50 subtype separation for TCGA-BRCA data.\label{tabscores}}%
\begin{tabular*}{\textwidth}{@{\extracolsep\fill}lcccccccccccc@{\extracolsep\fill}}
\hline%
\multirow{2}{*}{Method} & \multicolumn{3}{@{}c@{}}{SWISS score $\downarrow$} & \multicolumn{3}{@{}c@{}}{Davies-Bouldin index $\downarrow$} & \multicolumn{3}{@{}c@{}}{Silhouette index (in \%) $\uparrow$} & \multicolumn{3}{@{}c@{}}{Calinski-Harabasz index $\uparrow$}\\ \cline{2-13}%
 & mRNA      & DNA & miRNA & mRNA      & DNA & miRNA & mRNA      & DNA & miRNA & mRNA      & DNA & miRNA \\ \hline

All variables         & 0.841 & 0.842 & 0.918 & 3.74 & 4.93 & 4.95 & 1.92 & -3.66 & 1.16 & 49.62 & 37.54 & 23.62  \\
HSIC-SGCCA      & \textbf{0.295} & \textbf{0.454} & \textbf{0.752} & 4.60 & \textbf{3.38} & 4.19 & \textbf{10.39} & \textbf{1.66} & \textbf{3.50} & \textbf{627.75} & \textbf{315.81} & \textbf{86.73} \\ 
SA-KGCCA          & 0.844 & 0.989 & 0.992 & 3.79 & 99.25 & 68.99 & 1.50 & -14.30 & -8.59 & 48.68 & 2.85 & 2.11\\ 
TS-KGCCA         & \fbox{0.656} & \fbox{0.622} & 0.909 & \textbf{3.26} & \fbox{3.99} & 4.71 & \fbox{3.95} & 0.42 & \fbox{1.27} & \fbox{138.07} & \fbox{160.02} & 26.31 \\
DGCCA            & 0.847 & 0.884 & \fbox{0.882} & 4.43 & 4.48 & \textbf{4.02} & -0.70 & -1.02 & -0.57 & 39.54 & 49.13 & \fbox{49.59} \\ 
SUMCOR-SGCCA            & 0.841 & 0.875 & 0.887 & 3.74 & 4.93 & \fbox{4.11} & 1.91 & -3.66 & -0.27 & 49.59 & 37.53 & 33.35 \\
MAXVAR-SGCCA      & 0.826 & 0.858 & 0.915 & \fbox{3.57} & 4.27 & 4.86 & 1.53 & \fbox{0.54} & -0.02 & 55.26 & 43.64 & 24.34 \\
\hline
\end{tabular*}
 Note: $\downarrow$ means lower is better, $\uparrow$ means  higher is better, the best values are in \textbf{bold}, and the second-best values are in \fbox{\rule[0ex]{0pt}{1.2ex}~~~~}.\vspace{-0.3cm}
\end{table*}


To evaluate  the quality of variable selection, we  assess the ability of the selected variables to separate the PAM50 intrinsic breast cancer subtypes \citep{parker2009supervised}. We employ
four internal clustering validation metrics \citep{cabanski2010swiss,alma2014}: 
SWISS score, Davies-Bouldin index, Silhouette index,   and Calinski-Harabasz index.
These metrics are computed directly from the 
selected variables
and the predefined PAM50 subtype labels,
without applying any additional clustering or classification algorithms.
Table \ref{tabscores} reports the subtype separation results for each data view.
HSIC-SGCCA achieves the best scores in 10 out of 12 evaluation settings, with 8 of them substantially outperforming the second-best method.  It also records 1 third-best score (4.19, close to the best  4.02) and another score (4.60) that is near the best (3.26). TS-KGCCA attains 1 best score and 7 second-best scores.
These results indicate that capturing nonlinear dependencies leads to superior variable selection for subtype separation. Overall, HSIC-SGCCA exhibits the best~performance.





\begin{table}[!t]
\caption{Performance in XGBoost-AFT survival time prediction for TCGA-BRCA data. 
Results are given in average $\pm$ 1.96SE over 100 replications.\label{xgb-aft}}%
\scalebox{0.93}{\begin{tabular}{@{\extracolsep\fill}lccc@{\extracolsep\fill}}
\hline
 & hMAE  & hRMSE   & C-index  \\
Method & (in days) $\downarrow$ & (in days) $\downarrow$ & (in $\%$) $\uparrow$ \\
\hline
All variables & 65.11 $\pm$ 23.50  & 1226.08 $\pm$ 616.38 & 98.65 $\pm$ 0.27  \\
HSIC-SGCCA & \textbf{24.96} $\pm$ 2.26 & \textbf{298.77} $\pm$ 37.91  & \textbf{99.15} $\pm$ 0.10  \\
SA-KGCCA & \fbox{30.20} $\pm$ 3.54  & \fbox{409.75} $\pm$ 92.18 & \fbox{98.98} $\pm$ 0.17 \\
TS-KGCCA & 33.71 $\pm$ 5.88  & 479.40 $\pm$ 140.28  & 98.74 $\pm$ 0.31 \\
DGCCA & 44.25 $\pm$ 15.57  & 827.47 $\pm$ 458.15  & 98.93 $\pm$ 0.18  \\
SUMCOR-SGCCA & 57.92 $\pm$ 21.64  & 1066.84 $\pm$ 620.32  & 98.97 $\pm$ 0.21  \\
MAXVAR-SGCCA & 46.71 $\pm$ 20.71  & 910.99 $\pm$ 583.90  & 98.35 $\pm$ 0.36  \\
\hline
\end{tabular}}
 Note: $\downarrow$ means lower is better, $\uparrow$ means higher is better,  the best values are in \textbf{bold}, and the second-best values are in \fbox{\rule[0ex]{0pt}{1.2ex}~~~~}. \vspace{-0.3cm}
\end{table}

We also use the selected variables to predict
the survival time of breast cancer patients
from their initial pathological diagnosis.
This right-censored dataset includes 146 patients recorded as dead and 911 recorded as alive at their last follow-up.
We apply the state-of-the-art XGBoost-AFT model \citep{barnwal2022survival}, 
which 
effectively captures the nonlinear relationship between predictor features and survival time,
and handles high-dimensional data via $\ell_1$ and/or $\ell_2$ regularization.
For each GCCA method,
the predictor features used in XGBoost-AFT
consist of the selected variables from the three data views plus  8 clinical covariates: 
age at  diagnosis, race, ethnicity, pathologic stage, pathological subtype, PAM50 subtype, and two binary indicators for pharmaceutical treatment and radiation treatment.
We also consider the
method that uses all variables and the 8 covariates as predictor features.
Each method's XGBoost-AFT model is trained and tested over 100 replications, with the data randomly split into training and testing sets at a 4:1 ratio in each replication.
The prediction of survival time is evaluated
on each testing set 
using three metrics  \citep{qi2023survivaleval}:
hinge mean absolute error (hMAE),
hinge root mean squared error (hRMSE), and  concordance index (C-index).

 As shown in Table~\ref{xgb-aft},
our HSIC-SGCCA achieves the
lowest average hMAE and  hRMSE, both significantly 
outperforming the second-best values
with p-values of 0.008 and 0.022 from 
the t-test, respectively. 
Note that hMAE is more interpretable and robust than hRMSE.
HSIC-SGCCA attains an average hMAE of only 24.96 days, with a narrow 95\% confidence interval width of 4.52 days.
It also achieves the highest average C-index of 99.15\%, 
though all methods obtain average C-index values above 98.3\%. 
However,
C-index only measures a model's ability to correctly rank subjects by risk and does not assess the accuracy of predicted survival times.
Beyond  the top performance of   HSIC-SGCCA,
our SA-KGCCA ranks second in all three metrics, and 
our TS-KGCCA places third in  hMAE and hRMSE, further demonstrating the competitive performance of our proposed
methods.

\vspace{-0.35cm}
\section{Conclusion}\label{sec: conclusion}
In this paper, we propose three nonlinear SGCCA methods,  HSIC-SGCCA, SA-KGCCA, and TS-KGCCA, for
variable selection from multi-view high-dimensional data. 
These  methods are natural yet non-trivial 
extensions of 
SCCA-HSIC, SA-KCCA, and TS-KCCA
from two-view to multi-view settings, employing SUMCOR-like optimization criteria.
While SA-KGCCA and TS-KGCCA yield multi-convex optimization problems that we solve using BCD,
HSIC-SGCCA incorporates a necessary unit-variance constraint ignored by  SCCA-HSIC, resulting in an optimization problem that is neither convex nor multi-convex. 
We address this challenge by integrating BPL with LADMM.
The proposed HSIC-SGCCA achieves the best performance in variable selection in simulations,  as well as
in breast cancer subtype separation and
survival time prediction in the TCGA-BRCA data analysis.

Since the three proposed methods
are unsupervised,
they capture intrinsic associations
in multi-view data but do not
necessarily
ensure strong performance in downstream tasks, particularly supervised learning.
Unlike unsupervised (G)CCA methods, which are task-agnostic,
supervised (G)CCA 
methods \citep{jing2014intra,luo2016canonical}
are designed for specific tasks such as classification or regression
by  incorporating outcome information.
A natural future direction is  to develop supervised versions of our nonlinear SGCCA methods.
Another promising extension is to integrate structural information (e.g., group or graphical structures in genomic or brain data) into the penalty function, as this has been shown to enhance linear S(G)CCA methods for variable selection \citep{lin2013group,du2023adaptive}.

\vspace{-0.3cm}
\section*{Acknowledgements}
This work was supported in part through the NYU IT High Performance Computing resources, services, and staff expertise.
The real data analysis was based upon data generated by the TCGA Research Network: \url{https://www.cancer.gov/tcga}.

\vspace{-0.3cm}
\section*{Funding}
This work was partially supported by  Dr. Shu's
NYU GPH Goddard Award
and NYU GPH Research Support Grant.

\renewcommand\thesection{S\arabic{section}}
\setcounter{section}{0}

\renewcommand\theequation{S\arabic{equation}}
\setcounter{equation}{0}

\renewcommand\theremark{S\arabic{remark}}
\setcounter{remark}{0}

\renewcommand\thetable{S\arabic{table}}
\setcounter{table}{0}

\renewcommand\thefigure{S\arabic{figure}}
\setcounter{figure}{0}

\section*{Supplementary Material}

\section{Additional Related Work}

\subsection{S(G)CCA}
For CCA and GCCA,
the covariance matrix  
$\cov(\bd{x}_k,\bd{x}_\ell)$
in their
$
\cov(\bd{u}_k^\top\bd{x}_k,\bd{u}_\ell^\top\bd{x}_\ell)  
=\bd{u}_k^\top
\cov(\bd{x}_k,\bd{x}_\ell) \bd{u}_\ell
$ ($1\le k\le \ell\le K$)
is traditionally estimated by the
sample covariance matrix.
However, for high-dimensional data 
with $n=O(p_k)$,
the sample covariance matrix
is not a consistent estimator of the true covariance matrix \citep{MR950344}
due to  the accumulation of estimation errors over matrix entries.
To overcome the  curse
of high dimensionality,
SCCA 
and 
SGCCA methods (see articles cited in Section~\ref{sec: intro}, {\color{black}paragraph 3})
impose sparsity constraints on 
 canonical coefficient vectors  $\{\bd{u}_k\}_{k=1}^K$
to reduce the variable dimension, using
various penalties, optimization criteria, and algorithms.

For the SUMCOR GCCA in
\eqref{obj: SUMCOR},
\citet{witten2009extensions}
focus on 
the $\ell_1$ penalty 
$\{\|\bd{u}_k \|_1\le s_k\}_{k=1}^K$,
but use
the $\ell_2$-norm unit ball constraint
$\{\|\bd{u}_k \|_2\le 1\}_{k=1}^K$
instead of
the unit variance constraint 
to ease algorithm development. Their nonconvex but multi-convex
problem is solved using BCD
with a normalized soft-thresholding
for each $\bd{u}_k$ subproblem.
In contrast,
\citet{rodosthenous2020}
adopt
the convex relaxation 
$\{\var(\bd{u}_k^\top \bd{x}_k)\le 1\}_{k=1}^K$
of the unit variance constraint and~solve 
each $\bd{u}_k$ subproblem
via LADMM
with the $\ell_1$ or SCAD penalty. 
\citet{kanatsoulis2018structured} 
address SUMCOR GCCA
under the original unit variance constraint
using a penalty dual decomposition algorithm
for penalties with tractable proximal operators.

For the
MAXVAR GCCA in \eqref{obj: MAXVAR},
sparse variants
often remove 
the unit variance constraint
on $\bd{u}_k^\top \bd{x}_k$.
For instance,
\citet{fu2017scalable}
consider multiple 
sparsity-promoting
penalty options and
employ a BCD strategy that applies 
the proximal gradient method
to each $\bd{u}_k$ subproblem  
and the Procrustes projection 
to the $g$ subproblem.
\citet{9619966}
instead adopt the $\ell_0$ penalty 
and solve each $\bd{u}_k$ subproblem in BCD via
the Newton hard-thresholding pursuit.
\citet{lv2024sparse}
reformulate MAXVAR GCCA
as a linear system of equations, impose $\ell_1$ minimization 
on $\{\bd{u}_k\}_{k=1}^K$
to pursue sparsity,  and solve it using
 a distributed ADMM.

\subsection{K(G)CCA}\label{sec:kcca}

KCCA \citep{bach2002kernel,Fukumizu2007}
extends the linear CCA
to measure the nonlinear dependence between 
$\bd{x}_1\in \mathbb{R}^{p_1}$ and $\bd{x}_2\in \mathbb{R}^{p_2}$.
It 
maximizes the correlation 
between functions
in
their real-valued
RKHSs 
$\mathcal{H}_1$ and $\mathcal{H}_2$:
\be\label{eqn:KCCA}
\rho=\max_{\{f_k\in \mathcal{H}_{k}\}_{k=1}^2}
\cov(f_1(\bd{x}_1),f_2(\bd{x}_2))
~~\st~
\var(f_k(\bd{x}_k))=1.
\ee
Note that a real-valued RKHS associated with a Gaussian kernel 
accurately approximates the space of all functions with finite variance, making it a manageable surrogate for the latter, thereby facilitating easier computation
and analysis (see Section~\ref{sec: RKHS}).
However, the empirical kernel canonical correlation $\widehat{\rho}$ is 
always one and thus independent of data
when the 
kernel Gram matrices 
are invertible \citep{MR2249882}.
To address this issue,
the unit-variance constraint in~\eqref{eqn:KCCA}
is practically regularized
to be 
\be\label{eqn: KCCA reg constr}
\var(f_k(\bd{x}_k))+\varepsilon_k \| f_k\|_{\mathcal{H}_{k}}^2=1
\ee
with a small constant $\varepsilon_k>0$ \citep{Fukumizu2007}.

To enable variable selection,
SA-KCCA \citep{SAFKCCA}
assumes that
$f_k(\bd{x}_k)\in \mathcal{F}_k=\{\sum_{j=1}^{p_k}f_{kj}(\bd{x}_k^{[j]}):  \mathrm{E}[f_{kj}(\bd{x}_k^{[j]})]=0, f_{kj}\in \mathcal{H}_{kj}\}$
is a linear combination of individual
zero-mean functions 
$\{f_{kj}(\bd{x}_k^{[j]})\}_{j=1}^{p_k}$
with $f_{kj}$ in
a real-valued RKHS $\mathcal{H}_{kj}$
of $\bd{x}_k^{[j]}$, and 
 enforces sparsity  
on 
$\{f_{kj}(\bd{x}_k^{[j]})\}_{j=1}^{p_k}$
using a group Lasso penalty by solving:
\begin{align}\label{eqn: SA-KCCA}
\max_{\{f_k\in \mathcal{F}_k\}_{k=1}^2}
\cov(f_1(\bd{x}_1),f_2(\bd{x}_2))&\\
\st~
\var(f_k(\bd{x}_k))+\varepsilon_k \sum_{j=1}^{p_k}\| f_{kj}\|_{\mathcal{H}_{kj}}^2\le 1,~
\sum_{j=1}^{p_k}&
\sqrt{\mathrm{E}[f^2_{kj}(\bd{x}_k^{[j]})]}\le s_k.
\nonumber
\end{align}
The regularized 
variance 
inequality
in the first constraint
of \eqref{eqn: SA-KCCA}
is a convex relaxation
for its equality
counterpart similar to~\eqref{eqn: KCCA reg constr}.

Alternatively, 
TS-KCCA
\citep{yoshida2017sparse}
performs sparse multiple kernel learning in the first stage, followed by standard KCCA in the second stage.
 TS-KCCA
assumes 
 the kernel 
 $\kappa_k(\cdot, \bd{x}_k)$
 of 
the RKHS $\mathcal{H}_{k}$
for function $f_k$ in \eqref{eqn:KCCA}
as a liner combination
of the kernels 
 $\{\kappa_{kj}(\cdot, \bd{x}_k^{[j]})\}_{j=1}^{p_k}$
respectively from 
individual variables' RHKSs
$\{\mathcal{H}_{kj}\}_{j=1}^{p_k}$,
i.e.,
$\kappa_k(\cdot, \bd{x}_k)=\sum_{j=1}^{p_k} \bd{u}_k^{[j]}\kappa_{kj}(\cdot, \bd{x}_k^{[j]})$,
and selects variables via $\bd{u}_k$ based on HSIC:
\begin{align}\label{eqn: TS-KCCA}
&\max_{\{\bd{u}_k\in \mathbb{R}^{p_k}\}_{k=1}^2}\mathrm{HSIC}(\bd{x}_1,\bd{x}_2;\mathcal{H}_1,\mathcal{H}_2)\\
&\st~~ \bd{u}_k\ge 0,  
~~\|\bd{u}_k \|_2=1, ~ \|\bd{u}_k\|_1\le s_k.
\nonumber
\end{align}
Notably, the nonnegativity constraint is 
neither guaranteed in their algorithm
nor necessary  for inclusion  in the formulation.

To handle $K\ge 2$ data views,
\citet{tenenhaus2015kernel}
develop
a KGCCA method
using the SUMCOR
criterion \eqref{obj: SUMCOR}
with linear functions
 of  $\{\bd{x}_k\}_{k=1}^K$
replaced by 
functions in 
their 
real-valued RKHSs.
However, 
to the best of our knowledge,
a sparse version
of KGCCA is not available
in the existing literature. 

\subsection{DNN-based (G)CCA}\label{subsec: deepGCCA}
DNNs have high expressive power to approximate any continuous functions,
due to  universal approximation theorems \citep{gripenberg2003approximation}.
In objective~\eqref{eqn:KCCA},
DCCA \citep{andrew2013deep}
assumes the functions $\{f_k\}_{k=1}^2$ 
as DNNs 
instead of RKHS functions.
The DCCA variants, 
DCCAE \citep{wang2015deep}, DCCSAE \citep{li2020application} and DCCA-SCO \citep{DCCA-SCO},
utilize autoencoders to 
combine the DCCA objective
with  
reconstruction errors 
from each $f_k$ to $\bd{x}_k$.
Although DCCSAE and DCCA-SCO introduce sparsity, it is applied to hidden layer nodes of DNNs rather than the original variables of data views.
\citet{lindenbaum2021l0}
propose  $\ell_0$-DCCA, which induces sparsity 
 by
applying stochastic gates to $\{\bd{x}_k\}_{k=1}^K$ before feeding them into DCCA and penalizing the DCCA objective with the mean 
$\ell_0$ norm of the gates.

For $K\ge 2$ data views, DGCCA \citep{benton2017deep} extends DCCA using the MAXVAR criterion \eqref{obj: MAXVAR},
replacing $\bd{u}_k^\top(\bd{x}_k-\text{E}[\bd{x}_k])$ with 
$\bd{v}_k^\top \bd{f}_k(\bd{x}_k)$, where $\bd{f}_k$ is a vector-valued function modeled by a DNN.
Unlike $\bd{u}_k$,
the vector $\bd{v}_k$
cannot induce sparsity
in $\bd{x}_k$.
\citet{lindenbaum2021l0} briefly mention that $\ell_0$-DCCA can be extended to multi-view data
by replacing DCCA with DGCCA,
but no detailed implementation or analysis is provided.

\section{The Proposed SA-KGCCA and TS-KGCCA}
\subsection{SA-KGCCA}
We propose SA-KGCCA to extend the SA-KCCA \citep{SAFKCCA} in \eqref{eqn: SA-KCCA} to $K\ge 2$ data views using
a SUMCOR-like criterion:
\begin{align}\label{eqn: SA-KGCCA}
\max_{\{f_k\in \mathcal{F}_k\}_{k=1}^K}
\sum_{1\le s<t\le K}\cov(f_s(\bd{x}_s),f_t(\bd{x}_t))&\\
\st~
\var(f_k(\bd{x}_k))+\varepsilon_k \sum_{j=1}^{p_k}\| f_{kj}\|_{\mathcal{H}_{kj}}^2\le 1,
~\sum_{j=1}^{p_k}&
\sqrt{\mathrm{E}[f^2_{kj}(\bd{x}_k^{[j]})]}\le s_k,
\nonumber
\end{align}
where we use the same notation as defined above \eqref{eqn: SA-KCCA}.
Let $\mb{K}_{kj}\in\mathbb{R}^{n\times n}$
be the  Gram matrix
of kernel $\kappa_{kj}$
whose $(s,t)$-th entry 
$\mb{K}_{kj}^{[s,t]}=\kappa_{kj}((\bd{x}_k^{(s)})^{[j]},(\bd{x}_k^{(t)})^{[j]})$,
and define the centered Gram matrix as $\widetilde{\mb{K}}_{kj}=\mb{H}\mb{K}_{kj}\mb{H}$.
Following
\citet{SAFKCCA},
the empirical version
of \eqref{eqn: SA-KGCCA}
is 
\begin{subequations}\label{eqn: SA-KGCCA sample}
\begin{align}
&
\{\widehat{\bd{\alpha}}_{kj}\}_{j=1,k=1}^{p_k,K}
=\label{eqn 1: SA-KGCCA sample}\\
&\qquad\argmax_{\{\bd{\alpha}_{kj}\in\mathbb{R}^n\}_{j=1,k=1}^{p_k,K}}
\sum_{1\le s<t\le K}
\frac{1}{n} 
(\sum_{j=1}^{p_s}\widetilde{\mb{K}}_{sj}\bd{\alpha}_{sj})^\top
(\sum_{j=1}^{p_t} \widetilde{\mb{K}}_{t j}
\bd{\alpha}_{t j}
)\nonumber\\
&\st~~ 
\frac{1}{n} 
\|\sum_{j=1}^{p_k} \widetilde{\mb{K}}_{k j}\bd{\alpha}_{k j}\|_2^2
+
\varepsilon_k \sum_{j=1}^{p_k}
\bd{\alpha}_{kj}^\top\widetilde{\mb{K}}_{kj}\bd{\alpha}_{kj}\le 1    \label{eqn 2: SA-KGCCA sample}\\
&~\qquad\text{and}\quad
\sum_{j=1}^{p_k} \frac{1}{\sqrt{n}}\|
\widetilde{\mb{K}}_{kj} \bd{\alpha}_{kj}\|_2\le s_k.\label{eqn 3: SA-KGCCA sample}
\end{align}
\end{subequations}
Due to
$
 \frac{1}{\sqrt{n}} \|
\widetilde{\mb{K}}_{kj} \bd{\alpha}_{kj}\|_2\approx
\{\text{E}[f_{kj}^2(\bd{x}_k^{[j]})]\}^{1/2},
$
we 
 select variables 
 $\bd{x}_k^{[j]}$
 with
 $\frac{1}{\sqrt{n}} \|
\widetilde{\mb{K}}_{kj} \widehat{\bd{\alpha}}_{kj}\|_2\ne 0$.

The optimization problem \eqref{eqn: SA-KGCCA sample} is not convex
but is multi-convex,
as
it is convex
in each parameter block $\{\bd{\alpha}_{kj}\}_{j=1}^{p_k}$
when all
other parameter blocks
$\{\bd{\alpha}_{k'j}\}_{j=1}^{p_{k'}},k'\ne k$, are fixed.
To solve \eqref{eqn: SA-KGCCA sample}, 
we employ the BCD strategy that
alternately 
updates
$\{\bd{\alpha}_{kj}\}_{j=1}^{p_k}$ 
for $k=1,\dots, K$
by solving the convex subproblem:
\begin{align*}
&\max_{\{\bd{\alpha}_{kj}\}_{j=1}^{p_k}}
\frac{1}{n} 
(\sum_{j=1}^{p_k}\widetilde{\mb{K}}_{kj}\bd{\alpha}_{kj})^\top
\sum_{t=1:t\ne k}^K
(\sum_{j=1}^{p_t} \widetilde{\mb{K}}_{t j}
\bd{\alpha}_{t j}
)\\
&~~~~~~~~~~~\st\quad \eqref{eqn 2: SA-KGCCA sample}
\quad \text{and} \quad
\eqref{eqn 3: SA-KGCCA sample}.
\end{align*}
This subproblem
is 
a second-order cone program, which we solve using 
\texttt{SCS} \citep{ocpb16},
an ADMM-based solver available in the \texttt{CVXPY} Python library~\citep{CVXPY16}.
Following 
\citet{SAFKCCA} and \citet{bach2002kernel},
in our numerical studies, we 
use the Gaussian kernel
as $\kappa_{kj}$
with 
bandwidth $\sigma$ set to
the median of 
Euclidean distances between 
observations
of $\bd{x}_k^{[j]}$,
and set  $\varepsilon_k=0.02$ for all $k$.
{\color{black}The computational complexity of the algorithm without tuning is $O(RJn^2\sum_{k=1}^K p_k^2)$, where $R$ is the maximum number of outer iterations for BCD, where  all $\{\bd{\alpha}_{kj}\}_{j=1}^{p_k}$  for $k=1,\dots, K$ are updated in each iteration,
and $J$ is the maximum number of inner iterations used to run \texttt{SCS}
to solve the subproblem of $\{\bd{\alpha}_{kj}\}_{j=1}^{p_k}$.}

\subsection{TS-KGCCA}
We propose TS-KGCCA to extend the
TS-KCCA \citep{yoshida2017sparse} in \eqref{eqn: TS-KCCA} to $K\ge 2$ data views.
Our extension primarily modifies the first stage of TS-KCCA, with the second stage replacing KCCA  with KGCCA \citep{tenenhaus2015kernel}. The proposed first stage is formulated as the following optimization problem:
\begin{align}\label{eqn: TS-KGCCA}
&\max_{\{\bd{u}_k\in \mathbb{R}^{p_k}\}_{k=1}^K}
\sum_{1\le s<t\le K}
\mathrm{HSIC}(\bd{x}_s,\bd{x}_t;\mathcal{H}_s,\mathcal{H}_t)\\
&~~~~~~~~~~~~\st
~~\|\bd{u}_k \|_2\le 1, ~ \|\bd{u}_k\|_1\le s_k,  \nonumber
\end{align}
where the notation follows the definitions above \eqref{eqn: TS-KCCA}.
We discard
the nonnegativity  constraint $\bd{u}_k\ge 0$
used in \eqref{eqn: TS-KCCA}, 
as it is neither guaranteed in
the algorithm nor necessary for inclusion in the formulation of original TS-KCCA. 
We also replace
the unit $\ell_2$ norm constraint
$\|\bd{u}_k\|_2=1$ with
its convex relaxation
$\|\bd{u}_k\|_2\le 1$
to facilitate algorithm development.
Nonetheless,
 the resulting solution for $\bd{u}_k$  still satisfies
$\|\bd{u}_k\|_2=1$
due to the use of the
 normalized soft-thresholding method from \citet{witten2009penalized}.
 
The empirical version of the problem \eqref{eqn: TS-KGCCA} is
\be\label{TS-KGCCA sample}
\max_{\{\bd{u}_k\in \mathbb{R}^{p_k}\}_{k=1}^K}
\sum_{1\le s<t\le K} \bd{u}_s^\top
\mb{M}_{st} \bd{u}_t
~~\st
~~\|\bd{u}_k \|_2\le 1, ~ \|\bd{u}_k\|_1\le s_k,
\ee
where $\mb{M}_{st}\in \mathbb{R}^{p_s\times p_t}$
is a matrix with
 $(i,j)$-th entry
$
\mb{M}_{st}^{[i,j]}=
\tr(\mb{K}_{si}\mb{H}\mb{K}_{tj}\mb{H})/n^2
$
with $\mb{K}_{kj}$ ($1\le k\le K$, $1\le j\le  p_k$) defined below  \eqref{eqn: SA-KGCCA}.
The empirical  problem is not convex but is
multi-convex,
as it is convex 
in each parameter vector $\bd{u}_k$
when all other parameter vectors $\bd{u}_{k'},k'\ne k$
are fixed.
We solve it  using the  BCD strategy that
alternately updates 
$\bd{u}_k$ for $k=1,\dots, K$
by solving the convex subproblem:
\[
\max_{\bd{u}_k\in \mathbb{R}^{p_k}}
\bd{u}_k^\top
\sum_{t=1:t\ne k}^K \mb{M}_{kt} \bd{u}_t
~~\st
~~\|\bd{u}_k \|_2\le 1, ~ \|\bd{u}_k\|_1\le s_k,
\]
which
is  solved using the normalized soft-thresholding method from
\citet[see their Algorithm~3]{witten2009penalized}.
Same to SA-KGCCA, 
we use the Gaussian kernel
as $\kappa_{kj}$
with bandwidth $\sigma$ set 
to the  median inter-observation distance.
{\color{black}The computational complexity of this BCD algorithm 
without tuning is
$O((n^2 + R) \sum_{1\le s<t\le K}p_sp_t
 + RJ\sum_{k=1}^Kp_k )$,
  where $R$ is the maximum number of outer iterations for BCD, during which  all $\bd{u}_k$ for $k=1,\dots, K$ are updated in each iteration,
  and $J$ is the maximum number of
 inner iterations used to run  binary search
to
  find the  threshold for soft-thresholding 
  in the $\bd{u}_k$ subproblem. }

\section{Implementation Details}
  
 The computer code for all simulations and real-data analysis is available at \url{https://github.com/Rows21/NSGCCA}.
  
\subsection{Tuning and Multi-start for Proposed Methods}\label{subsec5}
We perform $M$-fold cross-validation
to tune the sparsity parameters $\{\lambda_k\}_{k=1}^K$ for HSIC-SGCCA
and $\{s_k\}_{k=1}^K$ for SA-KGCCA
and TS-KGCCA.
Specifically, for HSIC-SGCCA,
the $M$-fold cross-validation selects the optimal values
of tuning parameters $\{\lambda_k\}_{k=1}^K$
via grid search
over candidate values
$\{\{\lambda_k^{c_k}\}_{c_k=1}^{C_k}\}_{k=1}^K$
by maximizing
\begin{align*}
&\frac{1}{M} \sum_{m=1}^M 
  \sum_{1\le s<t\le K}      
   \widehat{\mathrm{HSIC}}(\{[\widehat{\bd{u}}_s^m(\lambda_s^{c_s})]^\top\bd{x}_s^{(i)},[\widehat{\bd{u}}_t^m(\lambda_t^{c_t})]^\top\bd{x}_t^{(i)}\}_{i\in S_m}) \\
&:=   \frac{1}{M} \sum_{m=1}^M 
  \sum_{1\le s<t\le K} 
\frac{1}{|S_m|}
\tr\Big(\mb{K}_s^m(\lambda_s^{c_s}) \mb{H} \mb{K}_t^m (\lambda_t^{c_t})\mb{H}\Big),
\end{align*}
where $S_m$ is the index set
of  the $m$-th subsample, 
$\{\widehat{\bd{u}}_k^m(\lambda_k^{c_k})\}_{k=1}^K$ are
estimates of $\{\bd{u}_k^*\}_{k=1}^K$ 
obtained using data
$\{\bd{x}_1^{(i)},\dots, \bd{x}_K^{(i)}\}_{i\notin S_m}$
with
tuning parameter values
$\{\lambda_k^{c_k}\}_{k=1}^K$,
and 
$
\mb{K}_k^m(\lambda_k^{c_k})\in\mathbb{R}^{|S_m|\times |S_m|}
$
is a kernel matrix
whose $(a,b)$-th entry
is $\exp(- \{ [\widehat{\bd{u}}_k^m(\lambda_k^{c_k})]^\top (\bd{x}_k^{(i_a)}- \bd{x}_k^{(i_b)}) \}^2  /2)
$
with $i_a$ the $a$-th largest value in
$S_m$.
Similarly, for SA-KGCCA and
TS-KGCCA, we select the optimal values of $\{s_k\}_{k=1}^K$ 
by maximizing their objective functions in \eqref{eqn 1: SA-KGCCA sample}
and \eqref{TS-KGCCA sample}, respectively. 

Since the optimization problems for the three proposed SGCCA methods are nonconvex, their algorithms based on 
BPL or BCD
may converge to a critical point  that is  not a global optimum. 
To alleviate this, we
adopt the routine multi-start strategy using multiple random initializations
for the parameters to be optimized \citep{mart2018handbook}.
For each initialization,
after determining the optimal tuning parameters via cross-validation,
we apply them to
the entire training dataset
to compute
the objective function.
The final solution is obtained from the initialization that yields the best objective value.

\subsection{Implementation Details of the Six GCCA methods}

The implementations are the same in 
 both simulations and real-data analysis for the six GCCA methods.

For our proposed 
HSIC-SGCCA, TS-KGCCA, and
SA-KGCCA,
we applied  10 random starts
in the multi-start strategy
and performed 5-fold cross-validation
for tuning, as described in 
Section~\ref{subsec5}.

For HSIC-SGCCA,
the sparsity parameter $\lambda_k$ was tuned within $\{10^{-4}, 10^{-3}, 10^{-2}, 10^{-1}\}$. The maximum number of
outer iterations ($R$) was 20
and that of inner iterations
($J$) was 50,
with an error tolerance of $5 \times 10^{-3}$.  

For TS-KGCCA,  the sparsity parameter $s_k$ was tuned over 10 evenly spaced values in $[1, \sqrt{p_k}]$. The error tolerance was $5 \times 10^{-2}$, with $R=10$ and $J=1000$.

For SA-KGCCA, the sparsity parameter $s_k$ was tuned over 10 evenly spaced values in $[1, \sqrt{p_k}]$.
The error tolerance was  
 $10^{-5}$, with $R=10$
 and $J=100$. 

For DGCCA \citep{benton2017deep}, we used three hidden layers (256, 512, and 128 units) and set the maximum number of epochs to 200. The learning rate was tuned within $\{10^{-4}, 10^{-3}, 10^{-2}, 10^{-1}\}$ via 5-fold cross-validation
based on the objective function (3) in  \citet{benton2017deep}
with $r=1$.
The code was obtained from \url{https://github.com/arminarj/deepgcca-pytorch}.

For SUMCOR-SGCCA \citep{kanatsoulis2018structured}, we tuned
the sparsity parameter $\lambda_k$ within $\{10^{-4}, 10^{-3}, 10^{-2}, 10^{-1}\}$ via 5-fold cross-validation based on the objective function \eqref{obj: SUMCOR}.
The error tolerance was $10^{-8}$, with
a maximum of 100 outer iterations and 
5 inner iterations.
The code was obtained  from \url{https://github.com/kelenlv/SGCCA_2023}.

For MAXVAR-SGCCA \citep{lv2024sparse}, we set the 
maximum number of 
outer iterations to 50
and that of 
inner iterations to 5,  $\beta_{\max} = 10^{4}$, $\rho = 1.0001$, and error tolerance $\epsilon_1 = \epsilon_2 = 10^{-5}$. The sparsity parameter $\delta$ was tuned within $\{10^{-4}, 10^{-3}, 10^{-2}, 10^{-1}\}$ using  5-fold cross-validation based on the 
unregularized objective function (2.4) in  \citet{lv2024sparse} with $\ell=1$. The code was obtained from \url{https://github.com/kelenlv/SGCCA_2023}.

All methods were run on 
an Intel Xeon Platinum 8268 CPU core (2.90GHz)
with 10GB memory 
for simulations
and  60GB memory for real-data analysis.
SUMCOR-SGCCA was implemented in MATLAB 2022a, while the other five methods were implemented in Python 3.8.

\subsection{Implementation Details of XGBoost-AFT}

We implemented  XGBoost-AFT 
using the  \texttt{xgboost} package \citep{chen2016xgboost} in Python (\url{https://xgboost.readthedocs.io/en/stable/tutorials/aft_survival_analysis.html}). 
For 
each GCCA method, the predictor features used in XGBoost-AFT consisted of the selected variables from the three-view TCGA-BRCA data along with 8 clinical covariates: age at diagnosis, race, ethnicity, pathologic stage, pathological subtype, PAM50 subtype, and two binary indicators for pharmaceutical treatment and radiation treatment. We also considered the method that uses all variables from the three views and the 8 covariates as predictor features. 
We standardized age at diagnosis to have zero mean and unit variance, while each of the seven categorical clinical covariates was converted into dummy variables, excluding the reference category.
Each method’s XGBoost-AFT model was trained and tested over 100 replications, with the data randomly split into training and testing sets at a 4:1 ratio in each replication. 
We performed 5-fold cross-validation on the training set for hyperparameter tuning, using grid search to optimize the learning rate in $\{0.01,0.1\}$, tree depth in $\{3,5,7\}$, $\ell_1$-regularization parameter in  $\{0.1,1,10\}$, and loss distribution scale in  $\{0.5,1.1,1.7\}$. We used
 negative log-likelihood 
(\texttt{aft-nloglik}) as the evaluation metric and the normal distribution (\texttt{normal}) as the AFT loss distribution. 
Once optimal hyperparameters were determined,
the final XGBoost-AFT model was trained on the full training set and evaluated on the testing set for survival time prediction.

\section{Additional Simulation Results}
Table~\ref{timesak} summarizes the runtime of SA-KGCCA in simulations.

\begin{table}[!t]
\caption{Runtime (average $\pm$ 1.96SE in seconds) for SA-KGCCA on simulation data based on 100 independent replications.
Each simulation setting is annotated with $(n,p,q)$.
\label{timesak}}%
\begin{tabular*}{\columnwidth}{@{\extracolsep\fill}lc|lc@{\extracolsep\fill}}
\hline
Linear & Time & Nonlinear & Time \\
\hline
(100, 30, 5) & $14.52\pm0.35$ & (100, 30, 5) & $14.45\pm1.20$ \\
(100, 50, 5) & $33.70\pm1.07$ & (100, 50, 5) & $27.42\pm0.33$ \\
(100, 100, 5) & $77.68\pm1.39$ & (100, 100, 5) & $79.58\pm1.23$\\ 
(100, 200, 5) & $213.49\pm8.30$ & (100, 200, 5) & $214.77\pm8.04$ \\
(200, 100, 5) & $448.11\pm15.83$ & (200, 100, 5) & $408.11\pm13.22$ \\ 
(400, 100, 5) & $1725.35\pm12.04$ & (400, 100, 5) & $1800.11\pm13.79$ \\
(100, 100, 10) & $78.96\pm0.84$ & (100, 100, 10) & $84.26\pm0.64$ \\ 
(100, 100, 20) & $95.02\pm3.28$  & (100, 100, 20) & $83.72\pm1.00$ \\ 
\hline
\end{tabular*}
\end{table}

\section{TCGA-BRCA Data Download and Preprocessing}
The R code used to download and preprocess the TCGA-BRCA data is available in 
\texttt{Data\_download\_preprocess.R}
at
\url{https://github.com/Rows21/NSGCCA}.
We obtained a three-view TCGA-BRCA dataset (GDC data release v41.0) for primary solid tumors in female patients using the \texttt{TCGAbiolinks} package  \citep[v2.34.0;][]{colaprico2016tcgabiolinks}. 
This dataset included mRNA expression, miRNA expression, and DNA methylation data.

Specifically, for mRNA expression data,
we downloaded RNA-seq gene expression counts generated by the STAR-Counts workflow, comprising 60,660 genes across 1098 samples. Low-expression genes were removed using the \texttt{filterByExpr} function in the \texttt{edgeR} package \citep[v4.4.1;][]{chen2025edger}  with default settings, retaining 18,213 genes. 
The counts were then normalized using the \texttt{DESeq2} package \citep[v1.46.0;][]{love2014moderated} and transformed using $\log_2(x+1)$.
Next, for  miRNA expression data, we retrieved  miRNA-seq counts for 1881 miRNAs, with 1081 samples overlapping those in the mRNA dataset. These miRNA counts were also normalized in \texttt{DESeq2} and 
$\log_2(x+1)$-transformed.
DNA methylation data were downloaded as 
$\beta$-values from two Illumina platforms: Human Methylation 450 (482,421 CpG sites, 781 samples) and Human Methylation 27 (27,578 CpG sites, 310 samples).
We merged them on the 25,978 CpG sites shared by the 1091 samples, excluded sites with more than 10\% missing data (yielding 23,495 sites), and imputed the remaining missing values using the \texttt{impute.knn} function in the \texttt{impute} package \citep[v1.80.0;][]{troyanskaya2001missing}. 
We then converted 
$\beta$-values to M-values and corrected for batch effects across the two platforms using the \texttt{ComBat} function in the \texttt{sva} package \cite[v3.54.0;][]{leek2012sva}, resulting in 1059 samples after intersecting with the mRNA and miRNA data.
Observed survival time was defined as days to last follow-up for living patients or days to death for deceased patients.
One patient with missing survival time and another with a negative time were removed, leaving 1057 samples. 

We further filtered the three-view data to focus on highly variable features. For the mRNA expression data, we retained 2596 genes whose standard deviation (SD) of 
$\log_2$-transformed counts exceeded 1.5.
For the miRNA expression data, 523 miRNAs were kept after discarding those with zero counts in more than half of the samples (i.e., $>$ 528 zeros).
For the DNA methylation data, we removed CpG sites with extremely low or high mean methylation levels ($|\text{E(M-value)}|>2$), retaining 6154 sites, and then kept 3077 sites whose SD of M-values was at least the median SD among those 6154 sites.
The final dataset thus consisted of 
$\log_2$-transformed mRNA expression counts for 2596 genes, $\log_2$-transformed miRNA expression counts for 523 miRNAs, and DNA methylation M-values for 3077 CpG sites, measured on 
a common set of 1057 primary solid tumor samples from 1057 female patients.

\bibliographystyle{abbrvnat2}
\bibliography{reference}

\end{document}